\documentclass{article}

\usepackage{PRIMEarxiv}

\usepackage[utf8]{inputenc} 
\usepackage[T1]{fontenc}    
\usepackage{hyperref}       
\usepackage{url}            
\usepackage{booktabs}       
\usepackage{amsfonts}       
\usepackage{nicefrac}       
\usepackage{microtype}      
\usepackage{lipsum}
\usepackage{fancyhdr}       
\usepackage{graphicx}       
\graphicspath{{media/}}     

\usepackage{times}
\usepackage{latexsym}

\usepackage[T1]{fontenc}

\usepackage[utf8]{inputenc}

\usepackage{microtype}

\usepackage{inconsolata}
\usepackage{microtype}
\usepackage{graphicx}

\usepackage{hyperref}

\usepackage[textsize=tiny]{todonotes}
\usepackage{amssymb}

\usepackage{natbib} 
\usepackage{adjustbox}
\usepackage{graphicx}
\usepackage{caption}
\usepackage{subcaption}

\usepackage{amsmath}
\usepackage{bm}
\usepackage{multirow}
\usepackage{xcolor}
\usepackage[utf8]{inputenc}
\usepackage{mathtools}
\usepackage{bbm}
\usepackage{physics}
\usepackage{float}
\usepackage{booktabs}
 \usepackage{array, makecell}
 \usepackage{algorithm}
 \usepackage{algorithmic}
\usepackage{tabularx}
\usepackage{amsthm}

\title{Suboptimal Shapley Value Explanations
}

\author{
  Xiaolei Lu\\
  \texttt{xiaoleilu2-c@my.cityu.edu.hk} \\
}

\begin{document}
\maketitle

\begin{abstract}
Deep Neural Networks (DNNs) have demonstrated strong capacity in supporting a wide variety of applications. Shapley value has emerged as a prominent tool to analyze feature importance to help people understand the inference process of deep neural models. Computing Shapley value function requires choosing a baseline to represent feature's missingness. However, existing random and conditional baselines could negatively influence the explanation. In this paper, by analyzing the suboptimality of different baselines, we identify the problematic baseline where the asymmetric interaction between $\bm{x}'_i$ (the replacement of the faithful influential feature) and other features has significant directional bias toward the model's output, and conclude that $p(y|\bm{x}'_i) = p(y)$ potentially minimizes the asymmetric interaction involving $\bm{x}'_i$. We further generalize the uninformativeness of $\bm{x}'_i$ toward the label space $L$ to avoid estimating $p(y)$ and design a simple uncertainty-based reweighting mechanism to accelerate the computation process. We conduct experiments on various NLP tasks and our quantitative analysis demonstrates the effectiveness of the proposed uncertainty-based reweighting mechanism. Furthermore, by measuring the consistency of explanations generated by explainable methods and human, we highlight the disparity between model inference and human understanding.
\end{abstract}


\section{Introduction}

Nowadays Deep Neural Networks (DNNs) have demonstrated impressive results on a range of tasks. For example, Language Models \cite{devlin2018bert,chowdhery2022palm,thoppilan2022lamda,openai2023gpt4} show strong capacity in the field of Natural Language Processing (NLP), Computer Vision (CV) and speech processing. Unlike traditional models (e.g. conditional random fields) that optimize weights on human interpretable features, deep neural models operate
like black-box models by applying multiple layers of non-linear transformation on the vector representations of input data, which fails to provide insights to understand the inference process of deep neural models.

Feature attribution methods \cite{kim2018interpretability} identify how much each feature contribute to the model's output, which could indicate how a deep model make decisions. Shapley value \cite{shapley1953value}, measuring the marginal contribution that a player makes upon joining the group by averaging over all possible permutations of players in the group, has been the prominent feature attribution to analyze feature importance toward model's output. For example, SHAP \cite{lundberg2017unified}, L-Shapley \cite{chen2018shapley} and WeightedSHAP \cite{kwon2022weightedshap}.

Computing Shapley value function requires choosing a baseline to represent feature's missingness. The most common solution is to fill in the missingness by randomly sampling from the dataset \cite{vstrumbelj2014explaining}. Another way is to generate features by conditioning on the observed features to replace the missing parts. However, \citet{frye2020shapley} and \citet{hooker2021unrestricted} argue that random sampling operation ignores inherent feature dependency. Also, it is difficult for conditional sampling based Shapley value to distinguish correlated feature with different sensitivity toward the model's output \cite{janzing2020feature, kumar2020problems}. \citet{watson2022rational} considered the baselines sampled from interventional distribution is optimal but it requires accessing to the underlying causal structure. 


In this paper, we study the faithfullness of Shapley-based model interpretation. Our contributions are summarized as follows:

\begin{itemize}

\item We analyze the suboptimality of different baselines for computing Shapley value in interpreting feature importance, and introduce asymmetric interaction to identify the problematic baseline where the interaction between $\bm{x}'_i$ (the replacement of the faithful influential feature) and other features has significant directional bias toward the model's output.

\item We propose that $p(y|\bm{x}'_i) = p(y)$ potentially minimizes the asymmetric interaction involving $\bm{x}'_i$. By generalizing the uninformativeness of $\bm{x}'_i$ toward the label space $L$ to avoid estimating $p(y)$, we design a simple uncertainty-based reweighting mechanism to accelerate the computation process.

\item We conduct quantitative analysis on various tasks to demonstrate the faithfulness of the proposed uncertainty-based reweighting mechanism, and measure the consistency of explanations generated by explainable methods and human to highlight the disparity between model inference and human understanding.
\end{itemize}

\section{Related work}

\subsection{Interpretability of deep models}

Different from white-box models (e.g. decision tree-based models) that are intrinsically interpretable, deep models operate like black-box models where the internal working mechanism are not easily understood. Existing methods for interpreting deep models could be categorized into two types: feature attribution that focuses on understanding how a fixed black-box model leads to a particular prediction, and instance attribution is to trace back to training data and study how a training point influence the prediction of a given instance. For example, Integrated Gradients \cite{sundararajan2017axiomatic} measures feature importance by computing the path integral of the gradients respect to each dimension of input. LIME \cite{ribeiro2016should} generates explanation by learning an inherently interpretable model locally on the instance being explained. Shapley value \cite{shapley1953value} that is derived from cooperative game theory treats each feature as a player and computes the marginal contribution of each feature toward the model's output. Regarding instance attribution, typical methods include influence function \cite{koh2017understanding} which attends to the final iteration of the training and TracIn \cite{pruthi2020estimating} that keeps tracing the whole training process.

\subsection{Shapley value}

Lloyd Shapley’s idea \cite{shapley1953value} is that players in the game should receive payments or shares proportional to their marginal contributions. \citet{vstrumbelj2014explaining} and \citet{lundberg2017unified} generalize Shapley value to measure feature importance toward the model's output by averaging marginal contribution of the feature being explained over all possible permutations among features. There are many variants of Shapley value to address efficiency and faithfulness. For example, to improve efficiency of Shapley value, \citet{chen2018shapley} proposes L-Shapley to explore local permutations and C-Shapley to compute valid connected nodes for structured data. \citet{kwon2022weightedshap} proposes WeightedSHAP to optimize weight of each marginal contribution under the user-defined utility function to emphasize influential features. 

Computing Shapley value function requires choosing a baseline to represent feature’s missingness. Random baseline \cite{lundberg2017unified,chen2018shapley, kwon2022weightedshap} has been widely used due to its immediate availability. Since random baseline ignores inherent feature dependency and thus could result in inconsistent context, \citet{frye2020shapley} and \citet{hooker2021unrestricted} address the importance of conditional baseline where the missingness is replaced with the features generated by conditioning on the observed features. However, \citet{janzing2020feature} and \citet{kumar2020problems} use simple linear model to demonstrate that conditional baseline based Shapley value fails to distinguish correlated feature with different sensitivity toward the model’s output. \citet{watson2022rational} considers that the optimal baseline should follow the interventional distribution but it requires accessing to the underlying causal structure. Generally for the above baselines none is more broadly applicable than the others, which motivates us to analyze suboptimality of these baselines for Shapley value on interpreting black-box models and find out the optimal baseline.

\section{Suboptimal baselines for Shapley value toward black-box model interpretation}

\subsection{Preliminaries}

In supervised setting, given an input $\bm{x}=\left ( \bm{x}_1,...,\bm{x}_n \right ) $ and a prediction model $f$, let $f_{y}(\cdot )$ denotes the model output probability on $y$, the Shapley value of $i_{th}$ feature in $\bm{x}$ for the model prediction $y$ is weighted and summed over all possible feature combinations as
\begin{equation}
\scalebox{0.95}{$
\phi(i)=\sum_{S\subseteq n\setminus {i}}\frac{(n-1-\left | S \right |)!\left | S \right |!}{n!}\left [ v(S\cup {i}) -v(S)\right ],$}
\end{equation}
where $S$ is the subset of feature indices. $v(S)$ is the value function that measures the change in prediction caused by observing certain subsets $S$ in the instance \cite{vstrumbelj2014explaining} and is defined as
\begin{equation}
v(S) =\mathbb{E}(f_{y}|\bm{x}_{S} \cup \bm{x}'_{\bar{S}}) -\mathbb{E}(f_{y}|\bm{x}'),
\end{equation}
where $\bm{x}'$ collectively denotes all possible instances and $ \mathbb{E}(f_y|\bm{x}')$ represents the model’s prediction averaged across these instances (this item could be excluded when computing the difference between $v(S\cup {i}) -v(S)$). $\bm{x}_{S} \cup \bm{x}'_{\bar{S}}$ refers to the instances that retain unchanged subset values $\bm{x}_S$. The contribution of the subset $S$ is determined by the difference between the effect of the subset $S$ and the average effect.

There are three common baselines \cite{watson2022rational} to fill in $\bm{x}'_{\bar{S}}$ that are described below.

\textbf{Random baseline}: $\bm{x}'_{\bar{S}}$ is drawn randomly from the distribution $p(\bm{x}'_{\bar{S}})$.

\textbf{Conditional baseline}: sample $\bm{x}'_{\bar{S}}$ by conditioning on $\bm{x}_{S}$.


\textbf{Interventional baseline}: $\bm{x}'_{\bar{S}}$ is generated by following the causal
data structure $p(\bm{x}'_{\bar{S}}|do(\bm{x}_S))$. Since accessing causal data structure is challenging we focus on discussing the above two baselines in this work.

\subsection{A motivation example}

\begin{figure}[h]
 \centering
 \begin{subfigure}[b]{0.5\textwidth}
     \centering
     \includegraphics[width=\textwidth,height=1.6in]{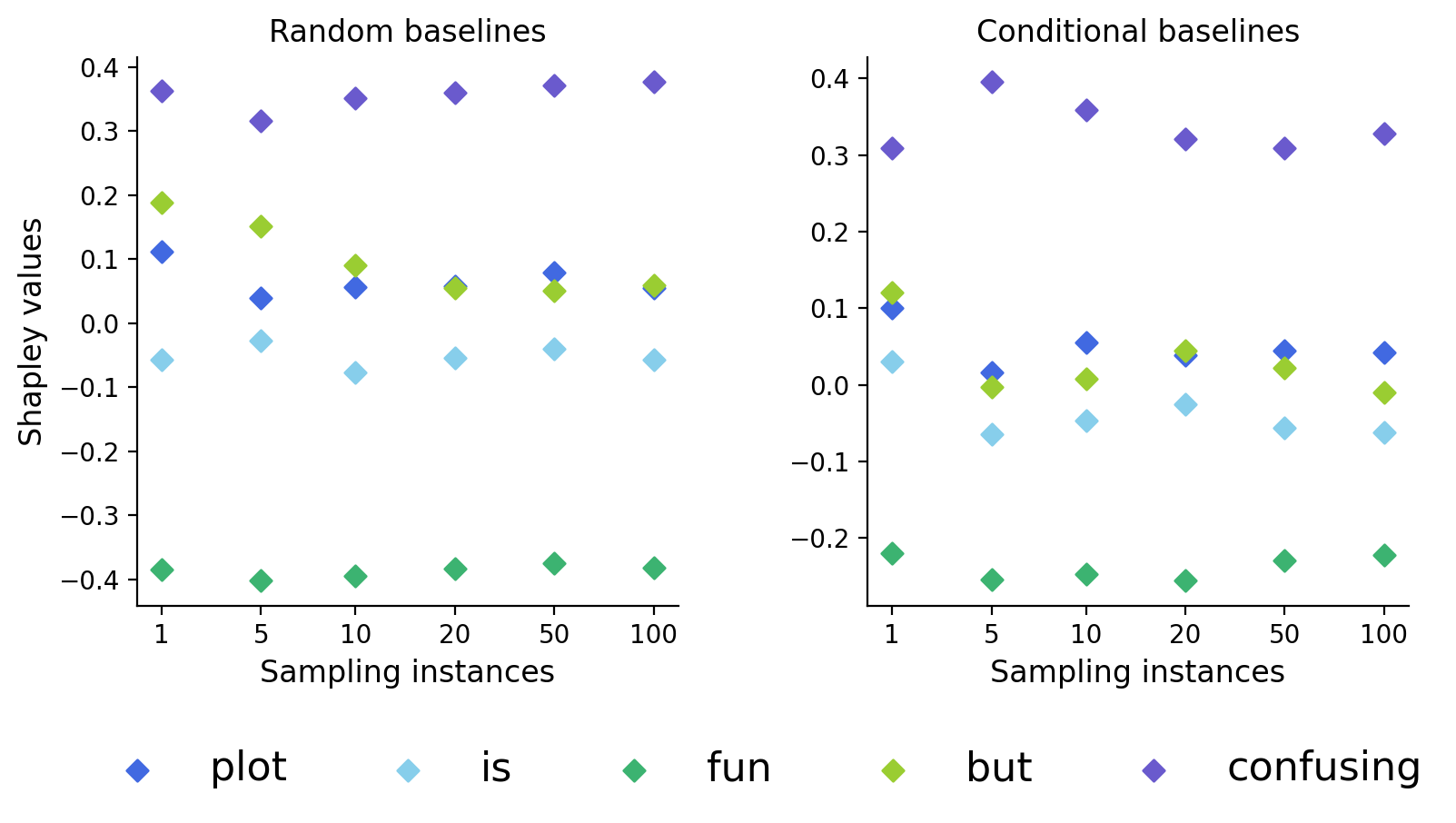}
     \caption{The ranking of Shapley values computed without the uncertainty-based reweighting mechanism.}
     \label{rank1}
 \end{subfigure}

 \begin{subfigure}[b]{0.5\textwidth}
     \centering
     \includegraphics[width=\textwidth,height=1.6in]{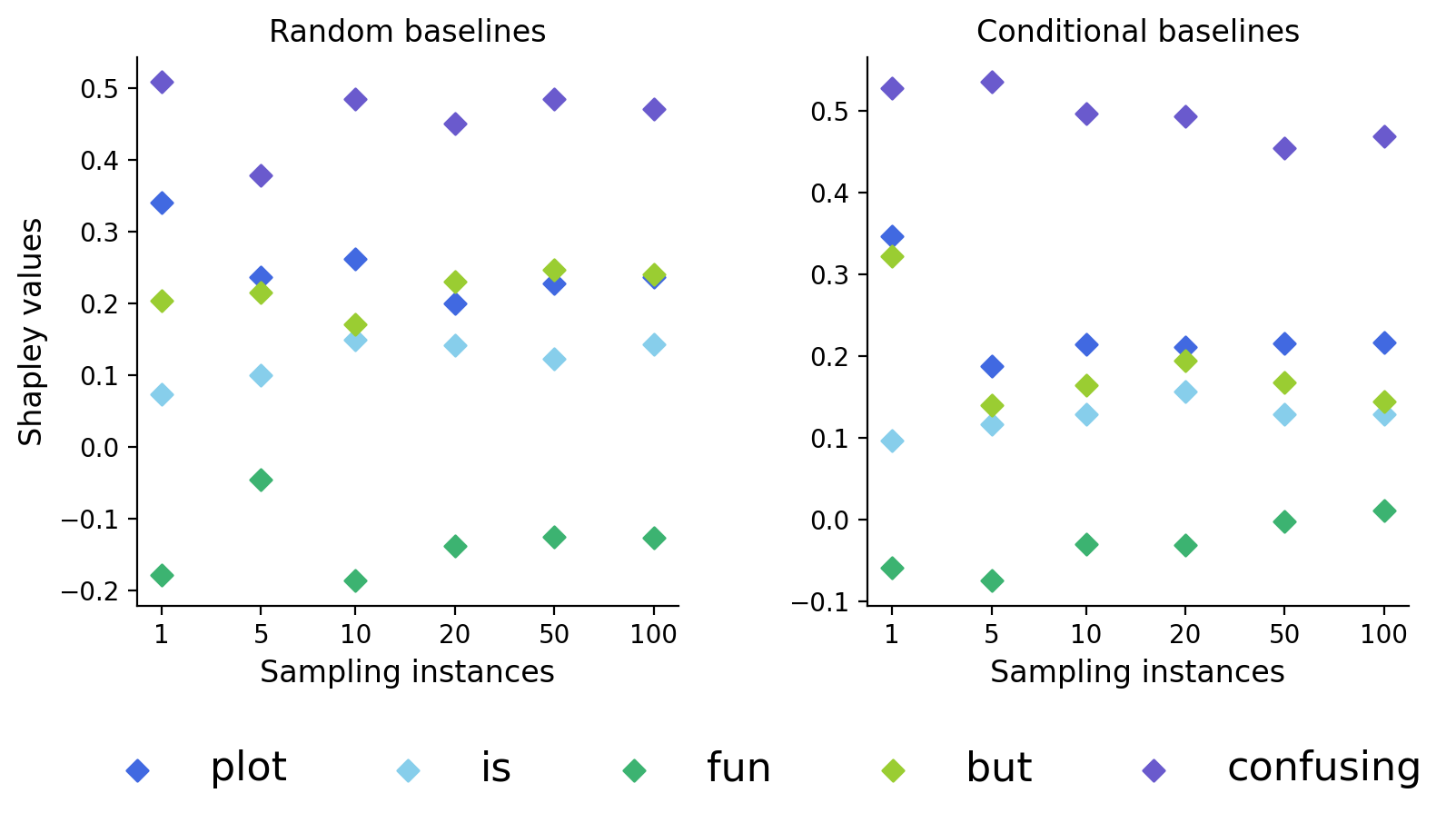}
     \caption{The ranking of Shapley values computed with the uncertainty-based reweighting mechanism.}
     \label{rank2}
 \end{subfigure}
 \caption{Comparison of the ranking of Shapley values without and with the uncertainty-based reweighting mechanism under different number of sampling instances. Sampling instances refer to the sampling size for computing $v(S)$.}
 \label{exprank}
\end{figure}

It could be trivial to show how the random and conditional baselines lead to incorrect interpretation on white-box models (analysis of a linear model is described in Appendix \ref{appendixa}). In interpreting deep models with Shapley value, we analyze an empirical example with a movie review ``\emph{plot is fun but confusing}'' that is predicted as ``Negative'' sentiment by our finetuned BERT classification model\footnote{The finetuned version is designed to predict this review as ``Negative" that is consistent with the ground-truth.}, the faithful feature contribution ranking\footnote{The faithful ranking is obtained by conducting quantitative analysis in Experimental section on the possible rankings provided by NLP researchers.} is [ ``\emph{confusing}" , ``\emph{plot}" , ``\emph{but}", ``\emph{is}", ``\emph{fun}" ]. Fig.\ref{exprank} shows the ranking of Shapley values computed by the random and conditional baselines with different number of sampling instances. 

\textbf{Random baseline}: As shown in Fig.\ref{rank1}, by increasing the number of sampling instances, ``\emph{confusing}" and ``\emph{fun}" are always identified as the most influential features while the ranks of ``\emph{plot}" and ``\emph{but}" are misleading in most cases. 

\textbf{Conditional baseline}: With smaller sampling size (i.e. $[1,20]$) that could ensure feature consistency via conditional sampling, we observe that in these scenarios ``\emph{confusing}" and ``\emph{fun}" are the most influential features but the contribution of ``\emph{plot}" and ``\emph{but}" could  be misinterpreted.

Given a fixed subset $S$, the contribution of $i_{th}$ feature is computed with
\begin{equation}
\begin{aligned}
\phi_S(i)= &v(S\cup i) - v(S)\\
=&\mathbb{E}(f_{y}|\bm{x}_{S\cup i} \cup \bm{x}'_{\bar{S}_1})-\mathbb{E}(f_{y}|\bm{x}_{S} \cup \bm{x}'_{\bar{S}_2}),\\
\end{aligned}
\end{equation}
where $\bar{S}_1\subseteq n\setminus {(S\cup i)}$ and  $\bar{S_2}\subseteq n\setminus {S}$.

For positively-contributing features, it is expected that without its participation $\mathbb{E}(f_y|\bm{x}_{S} \cup \bm{x}'_{\bar{S}_2})$ should be lower than $\mathbb{E}(f_y|\bm{x}_{S\cup i} \cup \bm{x}'_{\bar{S}_1})$. However, for highly influential positive features, under some random and conditional baselines $f_y(S\cup i)$ could be much lower than $f(S)$, which may results in lower $\phi_S(i)$ compared to less influential positive features. Similarly $\phi_S(i)$ for a highly influential negative feature could be higher than that of less influential negative features. 

Taking $\bm{x}_i$ = ``\emph{plot}" with $\bm{x}_{S} = $ ``\emph{is fun but}" as an example (more baselines with random sampling and conditional sampling are shown in Appendix \ref{appendixc}). With conditional sampling, $f_y(\emph{``practice is fun but brief"})=0.5147$ is much higher than $f_y(\emph{``plot is fun but brief"})=0.0347$ and $f_y(\emph{``script is fun but vague"})=0.8203$ exceeds $f_y(\emph{``plot is fun but vague"})=0.7113$. Obviously, compared with ``\emph{plot}", ``\emph{practice}" and ``\emph{script}" contribute more positively to $y$.


\subsection{Error analysis: asymmetric feature interaction view}

As described in Eq.(1), $\phi(\bm{x}_i)$ represents the average sum of the interactions between $\bm{x}_i$ and all possible subsets of $\bm{x}$ as
\begin{equation}
\phi(\bm{x}_i)=\sum_{\bm S(\bm{x}_i)\subseteq \bm{x}}\mathcal{I}(\bm{S}(\bm{x}_i)),
\end{equation}
where $\mathcal{I}(\bm{S}(\bm{x}_i))$ denotes the interaction within the subset $\bm{S}$ involving $\bm{x}_i$.

\begin{figure}
 \centering
 \begin{subfigure}[b]{0.4\textwidth}
     \centering
     \includegraphics[width=\textwidth,height=0.8in]{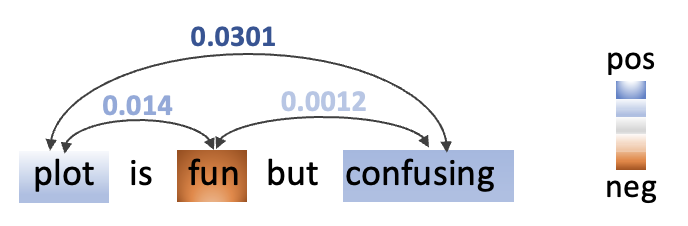}
     \caption{Pairwise symmetric feature interaction.}
     \label{sii}
 \end{subfigure}
 \begin{subfigure}[b]{0.4\textwidth}
     \centering
     \includegraphics[width=\textwidth,height=0.5in]{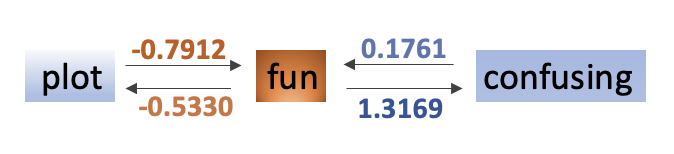}
     \caption{Pairwise asymmetric feature interaction.}
     \label{agg}
 \end{subfigure}
 \caption{Demonstrations of our defined asymmetric feature interaction and the symmetric feature interaction computed by the Shapley interaction index.}
 \label{sidd}
\end{figure}

Existing feature interaction attribution methods assume symmetric interaction where each feature contributes equally to the interaction score, which fails to capture imbalanced relationship across features toward model predictions. For example, as shown in Fig.\ref{sii}, we present pairwise interactions computed by Shapley interaction index\footnote{We employ random sampling with larger sampling size to approximate the faithful interactions, and the computed value may not be faithful due to inherent suboptimality.} \cite{grabisch1997k} among \{``plot", ``fun", ``confusing"\}. For positive influential ``confusing", interacting with either negative or positive features leads to a positive impact on the predicted outcome $y$, indicating that influence of ``confusing" can play a dominant role in some interactions. 

To further explore and quantify the variable influence of $\bm{x}_i$
within these interactions, we introduce the concept of asymmetric interaction.

 \noindent\textbf{Asymmetric interaction.} In the presence of the subset $\bm{x}_{T_2}$, the asymmetric interaction between two subsets $\bm{x}_{T_1}$ and $\bm{x}_{T_2}$ toward the predicted outcome $y$ is defined as 
\begin{equation}
\scalebox{0.85}{$
\begin{aligned}
&\phi(T_2\rightarrow T_1)\\
=& C_1\sum\limits_{T_2\subseteq S\subseteq n\setminus \left \{ T_1\right \}} \Delta (T_2,T_1,\bm{x}) -
C_2\sum\limits_{\substack{S_1, S_2\subseteq n\setminus \left \{ T_2\right \}\\
S_1\cap S_2 =\emptyset}}\Delta (T_2,S_2,\bm{x}),\\
\end{aligned}$}
\end{equation}
\begin{equation}
\Delta (T_2,T_1,\bm{x}) = I(y;\bm{x}_{T_1}|\bm{x}_S)-I(y;\bm{x}_{T_1}|\bm{x}_{S\setminus T_2}),\\
\end{equation}
\begin{equation}
\Delta (T_2,S_2,\bm{x})=I(y;\bm{x}_{T_2}|\bm{x}_{S_1}\cup \bm{x}_{S_2})-I(y;\bm{x}_{T_2}|\bm{x}_{S_1}),
\end{equation}
where $C_1=\frac{1}{2^{n-\left | T_1 \right |-\left | T_2 \right |}}$ and $C_2=\frac{1}{3^{n-\left | T_2 \right |}-2^{n-\left | T_2 \right |+1}+1}$ are the normalization terms. $I(\cdot)$ denotes the pointwise mutual information as
\begin{equation}
I(y;\bm{x}|\bm{x}')=\log \frac{p(y;\bm{x}|\bm{x}')}{p(y|\bm{x}')p(\bm{x}|\bm{x}')}.
\end{equation} 

$I(y;\bm{x}|\bm{x}')$ measures the amount of information that $\bm{x}$ contributes to $y$ when we already know $\bm{x}'$. $I(y;\bm{x}_{T_1}|\bm{x}_S)-I(y;\bm{x}_{T_1}|\bm{x}_{S\setminus T_2})$ quantifies the contribution of $\bm{x}_{T_1}$ toward $y$ in the presence of $\bm{x}_{T_2}$. Similarly, $I(y;\bm{x}_{T_2}|\bm{x}_{S_1}\cup \bm{x}_{S_2})-I(y;\bm{x}_{T_2}|\bm{x}_{S_1})$ measures the influence between $\bm{x}_{T_2}$ and $\bm{x}_{S_2}$ on $y$. Asymmetric interaction $\bm{x}_{T_2}\rightarrow \bm{x}_{T_1}$ is quantified by computing the difference between average interaction contribution involving $\bm{x}_{T_1}$ and $\bm{x}_{T_2}$, and the average interaction with $\bm{x}_{T_2}$. Fig.\ref{agg} demonstrates pairwise asymmetric interactions among \{``plot", ``fun"\} and  \{``fun", ``confusing"\}. We can observe that in the presence of different features, the corresponding interactions with ``fun" could have directional contribution toward the predicted outcome $y$.

 \noindent\textbf{Feature influence.} Given the asymmetric interaction graph $IG = (\bm{x}, \bm{E})$ where $\bm{x}$ is the set of features in a given instance and $\bm{E}$ is the set of all asymmetric interaction edges. Each asymmetric interaction edge is denoted as $\bm{e}=(\bm{e}_\mathrm{tail},\bm{e}_\mathrm{head})$.  $\text{IF}_{y}(\bm{x}_i)$, the influence of a feature $\bm{x}_i$ toward $y$, is measured by the corresponding head degree as 
\begin{equation}
\text{IF}_{y}(\bm{x}_i)=d(\bm{x}_i)=\sum_{\bm{e}\in \bm{E}}\phi(\bm{e})h_{\mathrm{head}}(\bm{x}_i,\bm{e}_{\mathrm{head}}),
\end{equation}
where $\phi(\bm{e})$ denotes the asymmetric interaction contribution along $\bm{e}$. $h_{\mathrm{head}}$ is defined as
\begin{equation}
h_{\mathrm{head}}(\bm{x}_i,\bm{e}_{\mathrm{head}}) =
    \begin{cases}
      1 & \text{if} \ \bm{x}_i \in \bm{e}_{\mathrm{head}},\\
      0 & \text{otherwise}.\\
    \end{cases} 
\end{equation}

In the asymmetric interaction graph, the higher the positive $\phi(\bm{e})$ of incoming edges, the more positive the influential feature. Conversely, incoming edges with lower negative $\phi(\bm{e})$ correspond to the negative influential feature. For example, as shown in Fig.\ref{agg}, the sum of $\phi(\bm{e})$ of incoming edges for ``\emph{fun}" is negative, which is consistent with its negative role in the faithful feature ranking. Generally, for a subset involving the highly influential feature, the interaction contribution is dominated by this feature. While the presence of less influential features do not diminish the asymmetric interaction of other features, their asymmetric interaction contributions tend to be negligible in the presence of influential features.

Given a fixed input instance $\bm{x}=\left ( \bm{x}_1,...,\bm{x}_n \right )$, the contribution of $\bm{x}_i$ toward the predicted outcome $y$ could be quantified by the sum of its individual importance and all asymmetric interactions involving $\bm{x}_i$ as 
\begin{equation}
\scalebox{0.95}{$
\begin{aligned}
 I(y;\bm{x}_i)+ \sum\limits_{\substack{T(\bm{x}_i) \subseteq n \\
 T'\subseteq S\subseteq n\setminus {T(\bm{x}_i)}}}\phi(T'\rightarrow T(\bm{x}_i)) + \phi(T(\bm{x}_i)\rightarrow T'),
\end{aligned}$}
\label{eqi}
\end{equation}
where $T(\bm{x}_i)$ denotes the subset involving $\bm{x}_i$.

Substituting $\bm{x}_i$ with $\bm{x}'_i$ resulting an increase or decrease in $f_y(\bm{x})$ implies that $\bm{x}'_i$ contributes more positively or negatively toward $y$ compared with $\bm{x}_i$. As the presence of less influential features only marginally affect the asymmetric interaction of other features, incorporating $\bm{x}'_i$ significantly contributes to 
\begin{equation}
\begin{aligned}
 I(y;\bm{x}'_i)  + \sum\limits_{\substack{T(\bm{x}'_i) \subseteq n \\
 T'\subseteq S\subseteq n\setminus {T(\bm{x}'_i)}}} \phi(T'\rightarrow T(\bm{x}'_i)).
\end{aligned}
\label{eqiiiii}
\end{equation}

\begin{figure}
 \centering
     \centering
     \includegraphics[width=0.4\textwidth]{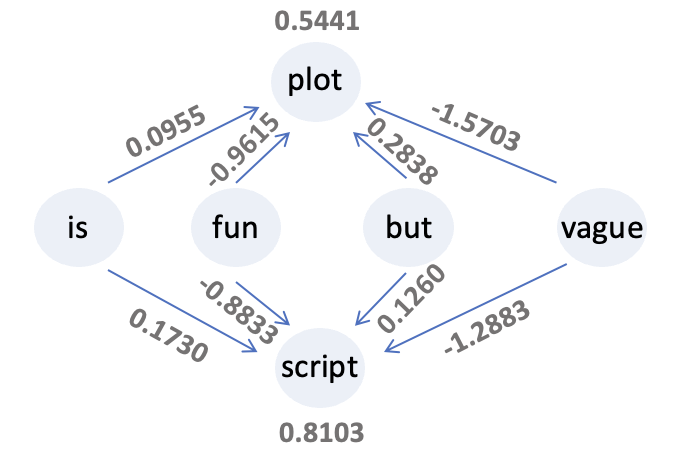}
     \caption{conditional baseline.}
     \label{asid_exp}
\end{figure}

As discussed in Section 3.2, by substituting ``\emph{plot}" with ``\emph{script}", $f_y(\emph{``plot is fun but vague"}) < f_y(\emph{``script is fun but vague"})$. Fig.\ref{asid_exp} demonstrates $I(y;\bm{x}_i)$ and pairwise asymmetric interaction $\phi(\bm{x}_j\rightarrow \bm{x}_i)$. We can observe that compared with ``\emph{plot}", ``\emph{script}" contributes less negatively toward $y$.

Therefore when computing $f_y(S)$, if the substitute of $\bm{x}_i$ substantially influences Eq.(\ref{eqiiiii}), $f_y(S)$ will be directional biased toward the predicted outcome $y$. In computing the Shapley value of faithful influential features with conditional baseline, this directional bias could result in a misinterpretation. As conditional sampling ensures the inherent consistency between features, the replacement of $\bm{x}_i$ in $\bm{x}_{S} \cup \bm{x}'_{\bar{S}_2}$ might be (more) influential as $\bm{x}_i$ toward $y$, then $\mathbb{E}(f_y|\bm{x}_{S} \cup \bm{x}'_{\bar{S}_2})$ would be larger than $ \mathbb{E}(f_y|\bm{x}_{S\cup i} \cup \bm{x}'_{\bar{S}_1})$ for ground-positive influential features and smaller for ground-negative influential features. Although increasing sampling size could mitigate these directional biases, it is computationally expensive for high-dimentional instances. Furthermore, larger sample sizes in conditional sampling are not recommended as these baselines ignore inherent feature dependency like random baseline.

\section{Proposed method}

To mitigate the directional bias toward $y$ generated by the substitute of $\bm{x}_i$ (i.e. $\bm{x}'_i$) in $\bm{x}_{S} \cup \bm{x}'_{\bar{S}_2}$ and then improve the importance of influential features, we should reduce the contribution of $\bm{x}'_i$ to Eq.(\ref{eqiiiii}) as
\begin{equation}
\min_{\bm{x}'_i\in C} \ \Big| I(y;\bm{x}'_i)  +  \sum\limits_{\substack{T(\bm{x}'_i) \subseteq n \\
 T'\subseteq S\subseteq n\setminus {T(\bm{x}'_i)}}}  \phi( T'\rightarrow T(\bm{x}'_i))\ \Big |,
\label{min}
\end{equation}
where $C$ denotes the set of possible features of $\bm{x}'_i$ to fill in $\bm{x}_{S} \cup \bm{x}'_{\bar{S}_2}$. 

Based on the triangle inequality (derivation details are provided in Appendix \ref{appendixb}), the above minimization problem is formulated with 
\begin{equation}
\scalebox{0.9}{$
\begin{aligned}
&\min_{\bm{x}'_i\in C} \ \Big| p(y|\bm{x}'_i)\ -p(y)\Big | + \sum\limits_{\substack{T(\bm{x}'_i) \subseteq n \\
 T' \subseteq S \subseteq n\setminus {T(\bm{x}'_i)}}} C_1 \ \Big| \Delta (T',T(\bm{x}'_i),\bm{x})\ \Big |,
\end{aligned}$}
\end{equation}
\begin{equation}
\scalebox{0.9}{$
\Delta (T',T(\bm{x}'_i),\bm{x})= \ \Big| I(y;\bm{x}_{T(\bm{x}'_i)}|\bm{x}_S)-I(y;\bm{x}_{T(\bm{x}'_i)}|\bm{x}_{S\setminus T'}) \ \Big |$}.
\end{equation}

For all possible $ T'$, if $I(y;\bm{x}_{T(\bm{x}'_i)}|\bm{x}_S) = I(y;\bm{x}_{T(\bm{x}'_i)}|\bm{x}_{S\setminus T'})$ holds true, it indicates that $T(\bm{x}'_i)$ consistently contributes to $y$ across different conditions. In the special case where $\bm{x}_{T(\bm{x}'_i)}$ is uninformative about $y$, we will have 
\begin{equation}
p(y;\bm{x}_{T(\bm{x}'_i)}|\bm{x}_S)=p(y|\bm{x}_S)p(\bm{x}_{T(\bm{x}'_i)}|\bm{x}_S),
\end{equation}
\begin{equation}
p(y;\bm{x}_{T(\bm{x}'_i)}|\bm{x}_{S\setminus T'})=p(y|\bm{x}_{S\setminus T'})p(\bm{x}_{T(\bm{x}'_i)}|\bm{x}_{S\setminus T'}),
\end{equation}
\begin{equation}
I(y;\bm{x}_{T(\bm{x}'_i)}|\bm{x}_S)=I(y;\bm{x}_{T(\bm{x}'_i)}|\bm{x}_{S\setminus T'})=0.
\end{equation}

When considering all possible combinations $\bm{x}_{T(\bm{x}'_i)}$ involving $\bm{x}_i$, if each combination is uninformative about $y$, it implies that $\bm{x}_i$ does not contribute to $y$. This is because if $\bm{x}_i$ is informative about $y$, there would be at least some configurations of $\bm{x}_{T(\bm{x}'_i)}$ contributing to $y$. Therefore given the special case that can minimize the asymmetric interaction of $\bm{x}'_i$, we conclude that 
\begin{equation}
p(y|\bm{x}'_i) = p(y),
\end{equation}
where $\bm{x}'_i$ is independent of $y$.

$p(y|\bm{x}'_i) = p(y)$ is the necessary but not sufficient condition for the uninformativeness of $\bm{x}_{T(\bm{x}'_i)}$ involving $\bm{x}'_i$ about $y$. Therefore $p(y|\bm{x}'_i) = p(y)$ leads to minimized $ I(y;\bm{x}'_i)$ and potentially achieves the special case where the asymmetric interaction involving $\bm{x}'_i$ is minimized. Minimizing Eq.(\ref{eqiiiii}) is conditionally equivalent \footnote{$p(y|\bm{x}'_i) = p(y)$ might not be the only optimal solution.} to 
\begin{equation}
\begin{aligned}
&\min_{\bm{x}'_i\in C} \ \Big| I(y;\bm{x}'_i)  +  \sum\limits_{\substack{T(\bm{x}'_i) \subseteq n \\
  T'\subseteq S\subseteq n\setminus {T(\bm{x}'_i)}}}  \phi( T' \rightarrow T(\bm{x}'_i))\ \Big |\\
\equiv & \min_{\bm{x}'_i\in C}\ \Big| p(y|\bm{x}'_i)-p(y)\ \Big |.
\end{aligned}
\end{equation}

As $p(y)$ depends on the size and characteristics of the data, we generalize the uninformativeness of $\bm{x}'_i$ toward the label space $L$ to avoid estimating $p(y)$ as
\begin{equation}
\max_{\bm{x}'_i \in C} -\sum_{j=1}^L p(y_j|\bm{x}'_{i})\log_2 p(y_j|\bm{x}'_{i}).
\end{equation}

By maximizing the entropy $H(Y|\bm{x}'_{i})$, the obtained $\bm{x}_{i}'^{*}$ is less certain regarding the label space, which is 
a practical and efficient alternative to minimize the difference between $p(y|\bm{x}'_i)$ and $p(y)$.

The greedy search for optimal $\bm{x}_{i}'^{*}$ in each baseline to compute $\mathbb{E}(f_y|\bm{x}_{S} \cup \bm{x}'_{\bar{S}_2})$ is computationally expensive, we design an uncertainty-based reweighting mechanism to accelerate the computation process as 
\begin{equation}
\scalebox{1}{$
\begin{aligned}
&\mathbb{E}(f_y|\bm{x}_{S} \cup \bm{x}'_{\bar{S}_2})\\
= &\mathbb{E}_{\bm{x}'_{\bar{S}_2}}\left [ \hat{H}(\bm{x}'_i)\cdot  f_y|\bm{x}_{S} \cup \bm{x}'_{\bar{S}_2}\right ],
\end{aligned}
$}
\end{equation}
where $\hat{H}(\bm{x}'_i)$ is the normalized entropy as
\begin{equation}
\hat{H}(\bm{x}'_i)=H(\bm{x}'_i)/ \log_2(L).
\end{equation}

In the reweighting mechanism higher weight $\hat{H}(\bm{x}'_i)$ is assigned to the sequence where 
$\bm{x}'_i$ is less certain toward $y$ and these baselines are encouraged to compute $\mathbb{E}(f_y|\bm{x}_{S} \cup \bm{x}'_{\bar{S}_2})$. Furthermore, with flexible generation of $\bm{x}'_{\bar{S}_2}$, this reweighting mechanism could be generalized to different baselines, which could improve the faithfulness of feature importance, especially in the scenario where the sampling size is limited. We present the results of applying the uncertainty-based reweighting mechanism to the random and conditional baselines on the motivation example in Fig.\ref{rank2} and it shows that the Shapley interpretations are consistent with the faithful ranking in most cases. 

\section{Experiments}
In this section, we evaluate the faithfulness of Shapley value with and without the uncertainty-based reweighting mechanism. By introducing the reweighting coefficient to the predicting expectation over random baselines (i.e. \textbf{random-uw}) and conditional baselines (i.e. \textbf{condition-uw}), we compare these Shapley interpretations with the original Shapley values computed by random and conditional baselines (denoted as \textbf{random} and \textbf{condition}). We adopt Shapley Sampling \cite{vstrumbelj2014explaining} to compute Shapley value (as summarized in Algorithm \ref{alg1} and \ref{alg2} in Appendix \ref{appendixd}). 

\subsection{Datasets and Models}
To effectively leverage language model for conditional sampling, we conduct experiments on various NLP tasks including sentiment analysis (SST-2 \cite{socher2013recursive} and Yelp-2 \cite{zhang2015character}), natural language inference (SNLI \cite{bowman2015large}), intent detection (SNIPS \cite{coucke2018snips}) and topic labeling (20Newsgroup\footnote{http://qwone.com/~jason/20Newsgroups/}), and  study the performance on BERT-base \cite{devlin2018bert} and RoBERTa-base \cite{liu2019roberta} models. Appendix \ref{appendixe} and \ref{appendixf} provide the configurations of finetuning pretrained BERT-base and RoBERTa-base models in these downstream tasks.

\subsection{Faithful evaluation metrics}

We choose three widely used faithful evaluation metrics and employ padding replacement operation. Since deletion-based evaluation yields similar results, we report the corresponding results in Appendix \ref{appendixde}. 

\textbf{Log-odds (LOR)} \cite{shrikumar2017learning}: average the difference of negative logarithmic probabilities on the predicted class over the test data before and after replacing the top $k\%$ influential words from the text sequence. The lower LOR, the more faithful feature importance ranking.

\textbf{Sufficiency (SF)} \cite{deyoung2019eraser}: measure whether important features identified by the explanation method are adequate to remain confidence on the original predictions. The lower SF, the more faithful feature importance ranking.

\textbf{Comprehensiveness (CM)} \cite{deyoung2019eraser}: evaluate if the features assigned lower weights are unnecessary for the predictions. The higher CM, the more faithful feature importance ranking.

\subsection{Quantitative Analysis}

To alleviate Out-of-Distribution (OOD) issue we choose $k$ within the range $\left [ 10,50 \right ]$. The sampling size of computing value function is set to $1000$ to enhance the robustness of feature importance ranking. The performance over RoBERTa architecture is reported in Appendix \ref{appendixro}.

Fig.\ref{er} demonstrates the evaluation performance over BERT architecture on different datasets. We can observe that \textbf{random-uw} generates more faithful explanations than other baselines. \textbf{random-uw} and \textbf{condition-uw} always perform better than the corresponding baselines without uncertainty-based reweighting. In SST-2, SNLI and Yelp-2, random-based baselines obtain better results compared with condition-based baselines while for SNIPS and 20Newsgroup \textbf{condition-uw} can outperform \textbf{random}. 

\begin{figure*}[h]
 \centering
 \begin{subfigure}[b]{\textwidth}
     \centering
     \includegraphics[width=\textwidth,height=0.9in]{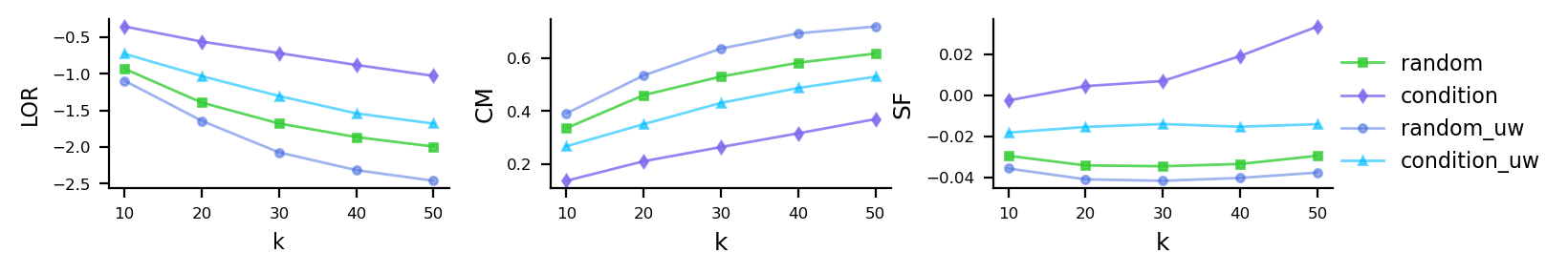}
     \caption{Evaluation performance over BERT on SST-2.}
     \label{fig1}
 \end{subfigure}
 \begin{subfigure}[b]{\textwidth}
     \centering
     \includegraphics[width=\textwidth,height=0.9in]{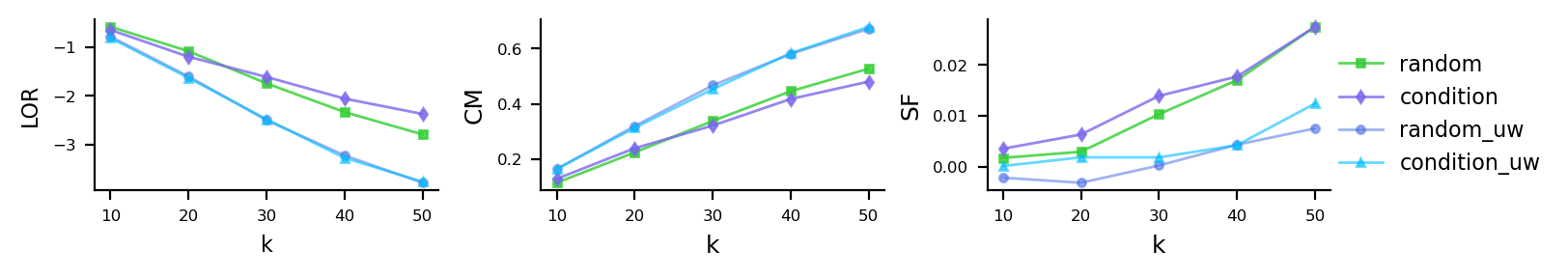}
     \caption{Evaluation performance over BERT on SNIPS.}
     \label{fig2}
 \end{subfigure}
 \begin{subfigure}[b]{\textwidth}
     \centering
     \includegraphics[width=\textwidth,height=0.9in]{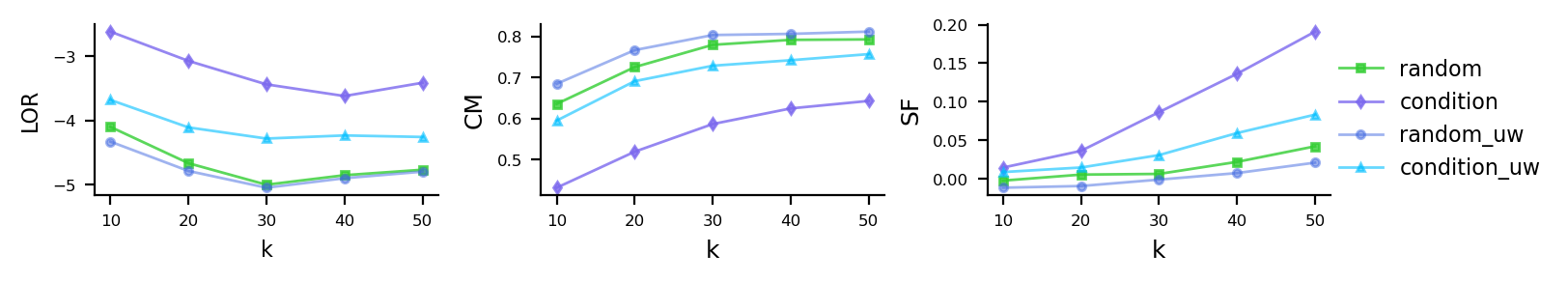}
     \caption{Evaluation performance over BERT on SNLI.}
     \label{fig3}
 \end{subfigure}
 \begin{subfigure}[b]{\textwidth}
     \centering
     \includegraphics[width=\textwidth,height=0.9in]{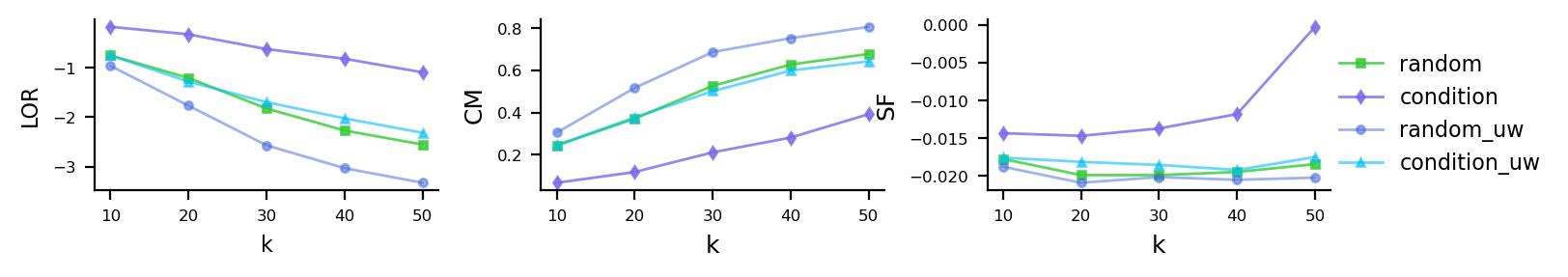}
     \caption{Evaluation performance over BERT on Yelp-2.}
     \label{fig4}
 \end{subfigure}
 \begin{subfigure}[b]{\textwidth}
     \centering
     \includegraphics[width=\textwidth,height=0.9in]{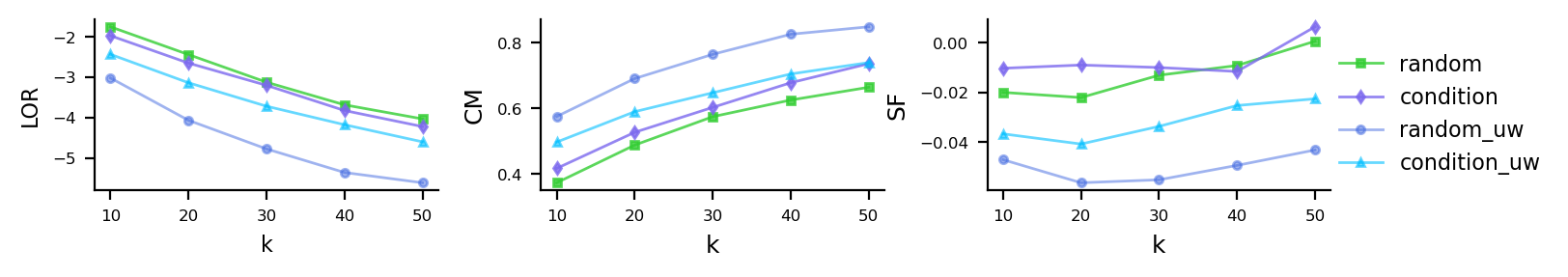}
     \caption{Evaluation performance over BERT on 20Newsgroup.}
     \label{fig2}
 \end{subfigure}
 \caption{Evaluation performances on the different datasets over BERT architecture.}
 \label{er}
\end{figure*}

Based on the above observations, introducing uncertainty-based reweighting to random and conditional baselines can improve the faithfulness of explanations. Furthermore, although random-based baselines are criticized for ignoring feature dependency, these baselines could generate informative local features without introducing additional dependency that tends to result in biased interaction \cite{janzing2020feature}. In particular, immediate availability of random baselines can greatly improve the efficiency of Shapley interpretation. We also show the robustness of Shapley Sampling method in Appendix \ref{rob}.

\subsection{Analysis of in-distribution baselines}

To address out-of-distribution baselines generated by random sampling or general pretrained language model \cite{kumar2020problems}, we pretrain BERT-base on these five datasets to ensure in-distribution sampling and evaluate in-distribution baselines (i.e. \textbf{condition-in} and  \textbf{condition-in-uw}) over BERT architecture. Fig.\ref{indi} shows the evaluation performance with $k=20$ (the results of Yelp-2 and SNIPS are reported in Appendix \ref{indii}). In-distribution baselines achieve significant improvement over their corresponding baselines without in-distribution sampling. In-distribution sampling does produce in-domain data. However, random sampling from training data also yields in-domain-like data. For example, for the input ``\emph{most new movies have a bright sheen}", given $\bm{x}_S$ = ``\emph{most new movies have}", in-distribution conditional sampling could generate ``\emph{most new movies have gone the back}" and random sampling yields ``\emph{most new movies have an extraordinary bore}" from the sequence ``\emph{a technical triumph and an extraordinary bore}".

\begin{figure*}
 \centering
 \begin{subfigure}[b]{0.3\textwidth}
     \centering
     \includegraphics[width=\textwidth]{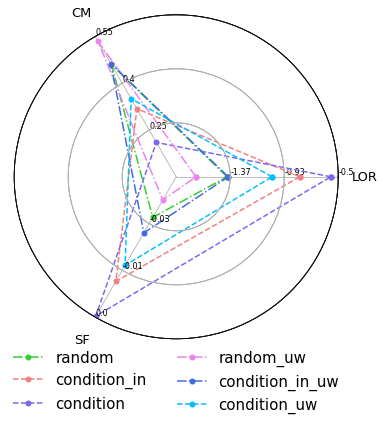}
     \caption{SST-2.}
     \label{si}
 \end{subfigure}
 \begin{subfigure}[b]{0.3\textwidth}
     \centering
     \includegraphics[width=\textwidth]{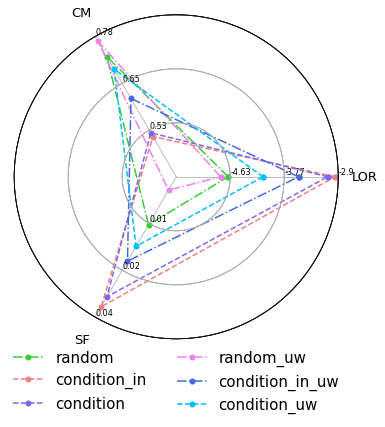}
     \caption{SNLI.}
     \label{ag}
 \end{subfigure}
 \begin{subfigure}[b]{0.3\textwidth}
     \centering
     \includegraphics[width=\textwidth]{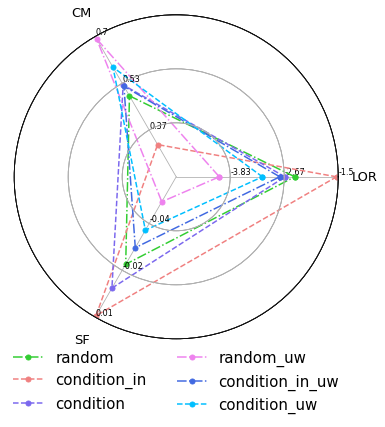}
     \caption{20Newsgroup.}
     \label{ag}
 \end{subfigure}
 \caption{Evaluation performance with $k=20$ in SST-2, SNLI and 20Newsgroup.}
 \label{indi}
\end{figure*}

\subsection{Human evaluation}
There are some research reports \cite{tornberg2023chatgpt,feng2023investigating} showing that large language models (LLMs), such as GPT-3.5 and GPT-4, can provide high-quality annotations like an excellent crowdsourced annotator does. Therefore we use GPT-4 to provide explanations toward the model's output (details of the prompt for generating explanation is given in Appendix \ref{he}). Using the rankings generated by GPT-4 as machine-generated version, we also provide human explanation grounded in human comprehension\footnote{Human evaluation is done by an NLP researcher and we will release these explanations in the public version.}.

To enable user-friendly human evaluation, we select 83 explained instances for BERT and 74 for RoBERTa on SST-2, and 86 explained instances for both BERT and RoBERTa on SNIPS. First, we employ Spearman’s Rank Correlation Coefficient to measure the correlation of feature importance ranking between GPT-4 (or human) and other baselines, and report the average results in Table \ref{thrank}. It could be observed that GPT-4 and human achieve great consistency, which demonstrates the quality of GPT-4 in human role.

\begin{table}
\caption{Rank correlation coefficient between GPT-4 (or human) and baselines over BERT architecture.}
\vskip 0.15in
\centering
\begin{tabular}{lcccc} 
\toprule
\multirow{2}{*}{Methods} & \multicolumn{2}{c}{SST-2}           & \multicolumn{2}{c}{SNIPS}            \\
                         & GPT-4            & Human            & GPT-4            & Human             \\ 
\hline
random                   & 0.086            & 0.097            & 0.175            & 0.129             \\
condition                & 0.071            & 0.115            & 0.213            & 0.165             \\
random\_uw               & 0.231            & 0.229            & 0.160            & 0.132             \\
condition\_uw            & 0.161            & 0.182            & 0.140            & 0.095             \\
GPT-4                    & \textbackslash{} & 0.710            & \textbackslash{} & 0.856             \\
Human                    & 0.710            & \textbackslash{} & 0.856            & \textbackslash{}  \\
\bottomrule
\end{tabular}
\label{thrank}
\end{table}

\begin{figure}
\centering
\includegraphics[width=1\linewidth,height = 1.3in ]{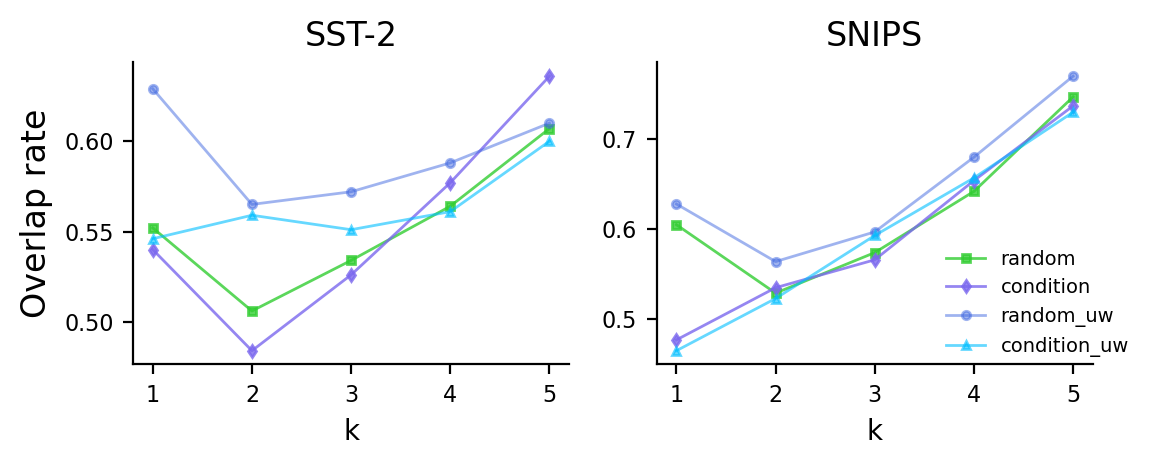}
\caption{Overlap rate of top $k\%$ influential features between human and baselines over BERT architecture on SST-2 and SNIPS.} 
\label{figop}
\end{figure}

Since feature importance ranking between GPT-4 (or human) and other baselines are weakly correlated, we further examine the overlap rate of influential features between GPT-4 (or human) and other baselines. As shown in Fig.\ref{figop}, the overlap rate is always within the range $\left [ 45\%,65\%\right ]$ under smaller $k$. By conducting quantitative analysis on GPT-4 and human explanations, as shown in Appendix \ref{he} Fig.\ref{hgss}, GPT-4 and human does not perform better than the compared baselines. It should be noted that there still exists a gap between model inference and human understanding \cite{keenan2023mind}.

\section{Conclusions}
In this paper we analyze the suboptimality of existing random and conditional baselines and identify the problematic baseline where the asymmetric interaction between $\bm{x}'_i$ (the replacement of the faithful influential feature) and other features has significant directional bias toward the model's output. We further design a simple uncertainty-based reweighting mechanism to mitigate these biased interactions. By evaluating different baselines on various tasks over BERT and RoBERTa architectures, quantitative analysis shows that our proposed uncertainty-based reweighting mechanism improves the faithfulness of Shapley-based interpretation. We measure the consistency of explanations generated by explainable methods, GPT-4 and human, which demonstrates the gap between model inference and human understanding.

\nocite{langley00}

\bibliographystyle{unsrtnat}
\bibliography{ref}

\newpage
\onecolumn
\appendix

\section{Suboptimal baselines for Shapley value
toward the linear model analysis}
\label{appendixa}

Consider a simple example $f(x_1,x_2) = x_1 + x_2$ where both $X_1$ and $X_2$ follow $\mathrm{Bernoulli} \ (\frac{1}{2})$ and 
\begin{equation}
p(x_1,x_2) =
    \begin{cases}
      \frac{1}{2} & \text{$x_1$=$x_2$},\\
      0 & \text{otherwise}.\\
    \end{cases} 
\end{equation}
For linear models, \citet{hooker2021unrestricted} shows that permutation-based importance methods associate variable importance with the magnitude of the corresponding coefficient when variable values are standardized. Therefore $X_1$ and $X_2$ should be equally informative in permutation-based approaches. Next we will introduce how is the Shapley values computed with different baselines.

\textbf{Random baseline}: $\mathbb{E}(f(x_1,X_2))$ and $\mathbb{E}(f(X_1,x_2))$ is computed as 
\begin{equation}
\begin{aligned}
\mathbb{E}(f(x_1,X_2))&=\mathbb{E}_{p(x_2)}(f(x_1,X_2))\\
&=p(x_2)f(x_1,x_2)_{|x_2=0} + p(x_2)f(x_1,x_2)_{|x_2=1}\\
&=\frac{1}{2}x_1 + \frac{1}{2}(x_1+1)\\
&=x_1+\frac{1}{2},
\end{aligned}
\end{equation}

\begin{equation}
\begin{aligned}
\mathbb{E}(f(X_1,x_2))&=\mathbb{E}_{p(x_1)}(f(X_1,x_2))\\
&=p(x_1)f(x_1,x_2)_{|x_1=0} + p(x_1)f(x_1,x_2)_{|x_1=1}\\
&=\frac{1}{2}x_2 + \frac{1}{2}(x_2+1)\\
&=x_2+\frac{1}{2},
\end{aligned}
\end{equation}

which yields 

\begin{equation}
\begin{aligned}
\phi_1= &\frac{1}{2}\left [ v(\emptyset \cup {1} )-v(\emptyset)+ v(2\cup {1} )-v(2)\right ]\\
=&\frac{1}{2}\left [ \mathbb{E}_{p(x_2)} (f(x_1,X_2))-\mathbb{E}_{p(x_1,x_2)} (f(X_1,X_2)) + f(x_1,x_2)-E_{p(x_1)} (f(X_1,x_2))\right ]\\
=&\frac{1}{2}\left [ x_1+\frac{1}{2}-1+x_1+x_2-x_2-\frac{1}{2}\right ]\\
=&x_1-\frac{1}{2},
\end{aligned}
\end{equation}

\begin{equation}
\begin{aligned}
\phi_2= &\frac{1}{2}\left [ v(\emptyset \cup {2} )-v(\emptyset)+ v(2\cup {1} )-v(1)\right ]\\
=&\frac{1}{2}\left [ \mathbb{E}_{p(x_2)} (f(x_1,X_2))-\mathbb{E}_{p(x_1,x_2)} (f(X_1,X_2)) + f(x_1,x_2)-\mathbb{E}_{p(x_1)} (f(X_1,x_2))\right ]\\
=&\frac{1}{2}\left [ x_2+\frac{1}{2}-1+x_1+x_2-x_1-\frac{1}{2}\right ]\\
=&x_2-\frac{1}{2},
\end{aligned}
\end{equation}
and $\phi_1 \neq \phi_2$.

\textbf{Conditional baseline}: $\mathbb{E}(f(x_1,X_2))$ and $\mathbb{E}(f(X_1,x_2))$ is computed as 
\begin{equation}
\begin{aligned}
\mathbb{E}(f(x_1,X_2))&=\mathbb{E}_{p(x_2|x_1)}(f(x_1,X_2))\\
&=p(x_2|x_1)f(x_1,x_2)_{|x_2=0} + p(x_2|x_1)f(x_1,x_2)_{|x_2=1}\\
&=2x_1\\
\end{aligned}
\end{equation}
\begin{equation}
\begin{aligned}
\mathbb{E}(f(X_1,x_2))&=\mathbb{E}_{p(x_1|x_2)}(f(X_1,x_2))\\
&=p(x_1|x_2)f(x_1,x_2)_{|x_2=0} + p(x_1|x_2)f(x_1,x_2)_{|x_2=1}\\
&=2x_2,
\end{aligned}
\end{equation}
where $p(x_1|x_2)=p(x_2|x_1)=1$ exclusively when $x_1=x_2$ and otherwise $p(x_1|x_2)=p(x_2|x_1)=0$. 

Considering two scenarios $x_1=x_2$ and $x_1\neq x_2$, we obtain
\begin{equation}
x_1=x_2, \phi_1=\phi_2=\frac{1}{2}(x_1+x_2-1),
\end{equation}
\begin{equation}
x_1=1,x_2=0, \phi_1=\frac{3}{2},\phi_2=-\frac{1}{2},
\end{equation}
\begin{equation}
x_1=0,x_2=1, \phi_1=-\frac{1}{2},\phi_2=\frac{3}{2},
\end{equation}
therefore with conditional baselines the generated explanations could be misleading.

\section{Derivation details of Eq.(\ref{min})}
\label{appendixb}

\begin{equation}
\begin{aligned}
&\ \Big| I(y;\bm{x}'_i)  +  \sum\limits_{\substack{T(\bm{x}'_i) \subseteq n \\
  T'\subseteq S\subseteq n\setminus {T(\bm{x}'_i)}}}  \phi( T' \rightarrow T(\bm{x}'_i))\ \Big | \\
 & \leq \ \Big| I(y;\bm{x}'_i)\Big |  +  \ \Big| \sum\limits_{\substack{T(\bm{x}'_i) \subseteq n \\
  T'\subseteq S\subseteq n\setminus {T(\bm{x}'_i)}}}  \phi( T' \rightarrow T(\bm{x}'_i))\ \Big | \\
 \end{aligned}
\end{equation}

As we focus on the items related to $\bm{x}'_i$, $\ \Big| \sum\limits_{\substack{T(\bm{x}'_i) \subseteq n \\
  T'\subseteq S \subseteq n\setminus {T(\bm{x}'_i)}}}  \phi( T' \rightarrow T(\bm{x}'_i))\ \Big |$ is minimized as 
\begin{equation}
\begin{aligned}
&\ \Big| \sum\limits_{\substack{T(\bm{x}'_i) \subseteq n \\
 T'\subseteq S \subseteq n\setminus {T(\bm{x}'_i)}}}  \phi(T'\rightarrow T(\bm{x}'_i))\ \Big | \\
&\leq \sum\limits_{\substack{T(\bm{x}'_i) \subseteq n \\
 T'\subseteq S \subseteq n\setminus {T(\bm{x}'_i)}}}  \ \Big|\phi(T'\rightarrow T(\bm{x}'_i))\ \Big | \\
&\leq  \sum\limits_{\substack{T(\bm{x}'_i) \subseteq n \\
 T' \subseteq S \subseteq n\setminus {T(\bm{x}'_i)}}} C_1 \ \Big| \Delta (T',T(\bm{x}'_i),\bm{x})\ \Big | \\
&\leq \scalebox{0.9}{$\sum\limits_{\substack{T(\bm{x}'_i) \subseteq n \\
 T'\subseteq S\subseteq n\setminus {T(\bm{x}'_i)}}}  C_1\ \Big| I(y;\bm{x}_{T(\bm{x}'_i)}|\bm{x}_S)-I(y;\bm{x}_{T(\bm{x}'_i)}|\bm{x}_{S\setminus T'}) \ \Big |$}\\
 \end{aligned} 
\end{equation}

\section{Demonstrations of the Shapley interpretation for the motivation example}
\label{appendixc}

\begin{figure}
 \centering
 \begin{subfigure}[b]{0.48\textwidth}
     \centering
     \includegraphics[width=\textwidth]{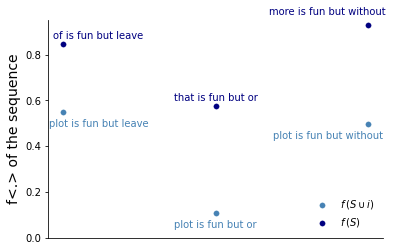}
     \caption{$f_y(S\cup i) < f_y(S)$ with random sampling.}
     \label{fig:ran}
 \end{subfigure}
 \hfill
 \begin{subfigure}[b]{0.48\textwidth}
     \centering
     \includegraphics[width=\textwidth]{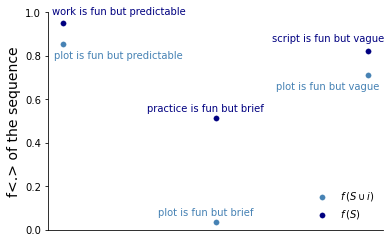}
     \caption{$f_y(S\cup i) < f_y(S)$ with conditional sampling.}
     \label{fig:con}
 \end{subfigure}
 \caption{The cases where $f_y(S\cup i) < f_y(S)$ with conditional sampling and random sampling.}
 \label{fig:plot}
\end{figure}

Taking $\bm{x}_i$ = ``\emph{plot}" with $\bm{x}_{S} = $ ``\emph{is fun but}" as an example, Fig.\ref{fig:plot} shows some cases where $f_y(S\cup i) < f_y(S)$ with conditional sampling and random sampling.

\section{Shapley Sampling algorithm}
\label{appendixd}
In the Experiment section, we use Shapley Sampling algorithm to estimate Shapley value with random and conditional baselines. The corresponding approximation algorithms are described as follows:
\begin{algorithm}
\caption{Shapley Sampling algorithm for approximating $\phi(i)$ (\textbf{random baseline}).}
\label{alg1}
\begin{algorithmic}
\STATE {\bfseries Input:} instance $\bm{x}\in \mathcal{X}$, prediction model $f$, sampling size $m$
\STATE Initialize $\phi(i)=0$
\FOR{$1$ {\bfseries to} $m$}
\STATE randomly select a permutation $\mathcal{O}\in \pi(n)$
\STATE randomly select a instance $\bm{x}'\in \mathcal{X}$
\STATE construct two instances: 
\STATE $\bm{x}_1 \leftarrow   \bm{x}_{\mathrm{precede} \ i_{} \ \mathrm{in}\ \mathcal{O}}  \cup   \bm{x}_i  \cup \bm{x}'_{\mathrm{succeed} \ i_{} \ \mathrm{in}\  \mathcal{O}}$ 
\STATE $\bm{x}_2 \leftarrow \bm{x}_{\mathrm{precede} \ i_{} \ \mathrm{in}\ \mathcal{O}} \cup   \bm{x}'_i  \cup \bm{x}'_{\mathrm{succeed} \ i_{} \ \mathrm{in}\  \mathcal{O}}$ 
\IF{uncertainty-based reweighting} 
\STATE  compute $\hat{H}(\bm{x}'_i)$
\STATE  $\phi(i) \leftarrow \phi(i)  +   f_y(\bm{x}_1) - \hat{H}(\bm{x}'_i)*f_y(\bm{x}_2)$
\ELSE
\STATE  $\phi(i) \leftarrow \phi(i)  +   f_y(\bm{x}_1) - f_y(\bm{x}_2)$
\ENDIF

\ENDFOR 
\STATE $\phi(i)\leftarrow\frac{\phi(i)}{m}$
\end{algorithmic}
\end{algorithm}

\begin{algorithm}
\caption{Shapley Sampling algorithm for approximating $\phi(i)$ (\textbf{conditional baseline}).}
\label{alg2}
\begin{algorithmic}
\STATE {\bfseries Input:} instance $\bm{x}\in \mathcal{X}$, prediction model $f$, sampling size $m$
\STATE Initialize $\phi(i)=0$
\FOR{$1$ {\bfseries to} $m$}
\STATE randomly select a permutation $\mathcal{O}\in \pi(n)$
\STATE construct two instances: 
\STATE \scalebox{0.95}{$\bm{x}'_{\mathrm{succeed} \ i \ \mathrm{in}\  \mathcal{O}} \sim p(\bm{x}'_{\mathrm{succeed} \ i \ \mathrm{in}\  \mathcal{O}}|\bm{x}_{\mathrm{precede} \ i \ \mathrm{in}\  \mathcal{O}} \cup \bm{x}_i )$}

\STATE $\bm{x}_1 \leftarrow   \bm{x}_{\mathrm{precede} \ i_{} \ \mathrm{in}\ \mathcal{O}}  \cup  \bm{x}_i  \cup \bm{x}'_{\mathrm{succeed} \ i_{} \ \mathrm{in}\  \mathcal{O}}$ 

\STATE \scalebox{0.95}{$\bm{x}'_{\mathrm{succeed} \ i \ \mathrm{in}\  \mathcal{O}} \cup  \bm{x}'_{i}\sim p(\bm{x}'_{\mathrm{succeed} \ i \ \mathrm{in}\  \mathcal{O}}\cup \bm{x}'_i|\bm{x}_{\mathrm{precede} \ i \ \mathrm{in}\  \mathcal{O}})$}
\STATE $\bm{x}_2 \leftarrow \bm{x}_{\mathrm{precede} \ i_{} \ \mathrm{in}\ \mathcal{O}} \cup   \bm{x}'_i  \cup \bm{x}'_{\mathrm{succeed} \ i_{} \ \mathrm{in}\  \mathcal{O}}$ 

\IF{uncertainty-based reweighting} 
\STATE  compute $\hat{H}(\bm{x}'_i)$
\STATE  $\phi(i) \leftarrow \phi(i)  +   f_y(\bm{x}_1) - \hat{H}(\bm{x}'_i)*f_y(\bm{x}_2)$
\ELSE
\STATE  $\phi(i) \leftarrow \phi(i)  +   f_y(\bm{x}_1) - f_y(\bm{x}_2)$
\ENDIF

\ENDFOR 
\STATE $\phi(i)\leftarrow\frac{\phi(i)}{m}$
\end{algorithmic}
\end{algorithm}

\section{Details of the tasks and datasets}
\label{appendixe}
\renewcommand{\arraystretch}{1.5}
\begin{table}[H]
\caption{Summary of the selected datasets (we preprocess the datasets to remove the sequences containing one word and the punctuations at the end of the sequences), where Explained set is randomly sampled from the Test set to avoid high computational complexity and Average sequence length denotes the average sequence length of the instances in Explained set (the length of Yelp-2 explained set is controlled within 50 and we will provide these Explained sets in the public version).}
\vskip 0.15in
\centering
\label{datasum}
\begin{adjustbox}{width=1\textwidth}
\begin{tabular}{lcccccc} 
\toprule
Datasets    & Category & Train set & Test set & Explained set & Label set & Average sequence length \\  
\hline 
SST-2       & sentimental analysis         &     6899                          &  1819                           &   1000            &   2        &  19.24                                            \\
SNIPS       &  intent detection        &   13082                            &         700                    &   700           &    7       &    9.08                                          \\
SNLI        &  natural language inference        &  54936                            &   7824                            &   1000           &     3      &  21.22                                            \\
Yelp-2      &   sentimental analysis      &  559694
      &  37981        &  200             &   2        &    32.12                   \\
20Newsgroup &    topic labeling      &   10663                             &     7019                       &   300            &   20        &      143.28                                        \\
\bottomrule
\end{tabular}
\end{adjustbox}
\end{table}

\section{Configurations for finetuning deep models}
\label{appendixf}
We use AdamW optimizer with weight decay $0.001$ and start with learning rate of 2e-5 to tune pretrained BERT-base-uncased and RoBERTa-base models. For the setting of epochs and batch size, SST-2 at 10 epochs with a batch size of 32, Yelp-2 also at 10/32, SNLI and SNIPS both at 10/64, and NG at 20/64, ensuring good model performance for each task. The corresponding performance is reported in Table \ref{appendixff}.

\begin{table}[H]
\caption{Task performance with finetuning BERT and RoBERTa.}
\vskip 0.15in
\centering
\begin{tabular}{lccccc} 
\toprule
Models  & SST-2 & Yelp-2 & SNLI  & SNIPS & 20Newsgroup  \\ 
\hline
BERT    & 90.49 & 96.20  & 89.44 & 97.71 & 74.48        \\
RoBERTa & 94.56 & 96.87  & 90.19 & 97.85 & 73.37        \\
\bottomrule
\end{tabular}
\label{appendixff}
\end{table}

\section{Details of the evaluation metrics}
\label{appendixg}
\textbf{Log-odds (LOR)} \cite{shrikumar2017learning}: average the difference of negative logarithmic probabilities on the predicted class over the test
data before and after replacing the top $k\%$ influential words from the text sequence.

\begin{equation}
    \mathrm{LOR}(k) = \frac{1}{N}\sum_{i=1}^N \log \frac{f(\bm{x'}_i)}{f(\bm{x}_i)},
\end{equation}
where $\bm{x'}_i$ is obtained by replacing the $k\%$ top-scored words from $\bm{x}_i$. The lower LOR, the more faithful feature importance ranking.

\textbf{Sufficiency (SF)} \cite{deyoung2019eraser}: measure whether important features identified by the explanation method are adequate to remain confidence on the original predictions. 
\begin{equation}
    \mathrm{SF}(k) = \frac{1}{N}\sum_{i=1}^N f(\bm{x}_i)-f(\bm{x}_{i,k\%}),
\end{equation}
where $\bm{x}_{i,k\%}$ is obtained by replacing non-top $k\%$ influential elements in $\bm{x}_i$. The lower SF, the more faithful feature importance ranking.

\textbf{Comprehensiveness (CM)} \cite{deyoung2019eraser}: evaluate if the features assigned lower weights are unnecessary for the predictions.
\begin{equation}
   \mathrm{CM}(k) = \frac{1}{N}\sum_{i=1}^N f(\bm{x}_i)-f(\bm{x}_{i}\backslash \bm{x}_{i,k\%}),
\end{equation}
where $\bm{x}_{i}\backslash \bm{x}_{i,k\%}$ is obtained by replacing top $k\%$ influential elements in $\bm{x}_i$. The higher CM, the more faithful feature importance ranking.

\section{Performance over BERT architecture with deletion operation}
\label{appendixde}

\begin{figure*}
 \centering
 \begin{subfigure}[b]{\textwidth}
     \centering
     \includegraphics[width=\textwidth,height=1in]{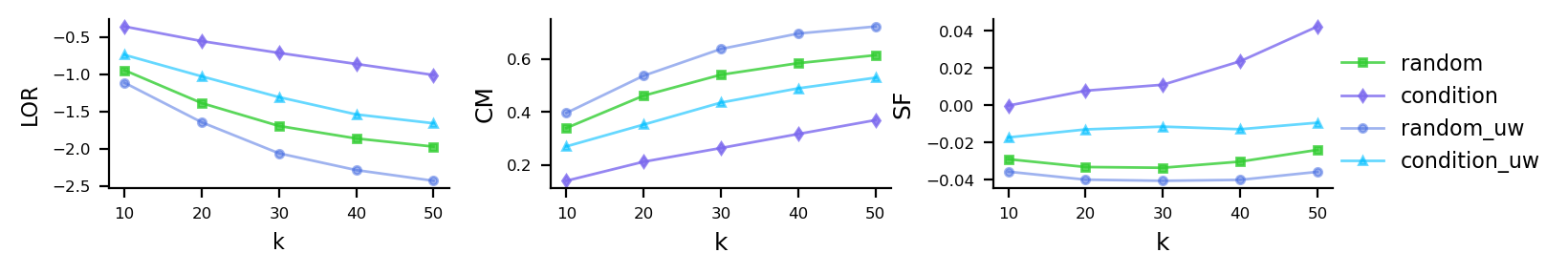}
     \caption{Evaluation performance over BERT on SST-2.}
     \label{fig1}
 \end{subfigure}
 \begin{subfigure}[b]{\textwidth}
     \centering
     \includegraphics[width=\textwidth,height=1in]{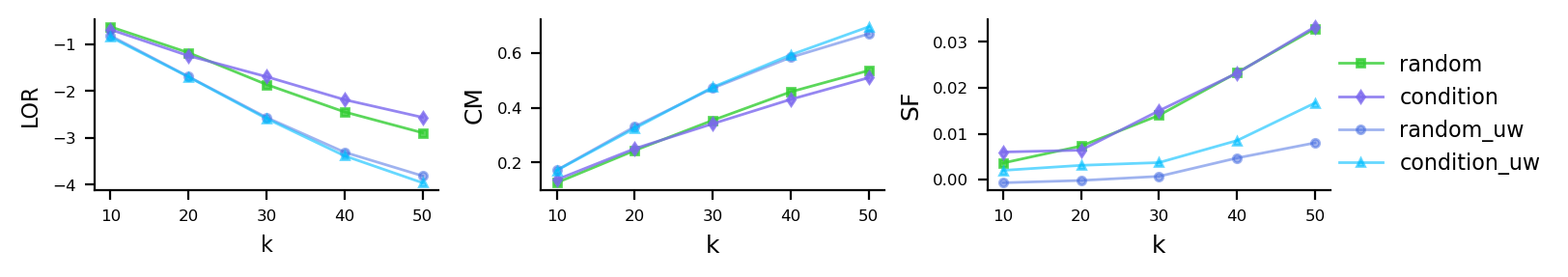}
     \caption{Evaluation performance over BERT on SNIPS.}
     \label{fig2}
 \end{subfigure}
 \begin{subfigure}[b]{\textwidth}
     \centering
     \includegraphics[width=\textwidth,height=1in]{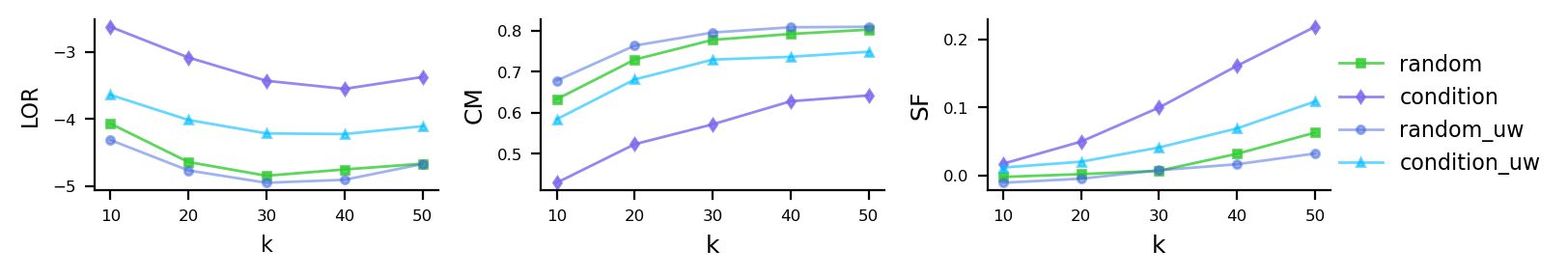}
     \caption{Evaluation performance over BERT on SNLI.}
     \label{fig3}
 \end{subfigure}
 \begin{subfigure}[b]{\textwidth}
     \centering
     \includegraphics[width=\textwidth,height=1in]{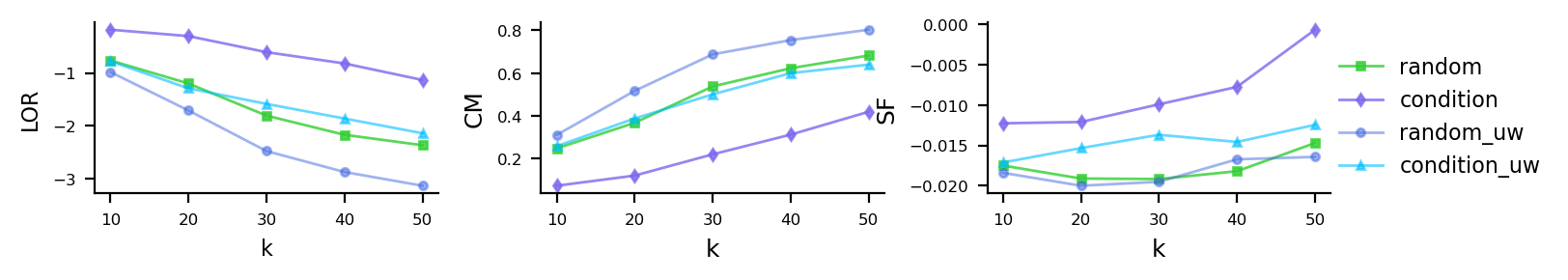}
     \caption{Evaluation performance over BERT on Yelp-2.}
     \label{fig4}
 \end{subfigure}
 
 \begin{subfigure}[b]{\textwidth}
     \centering
     \includegraphics[width=\textwidth,height=1in]{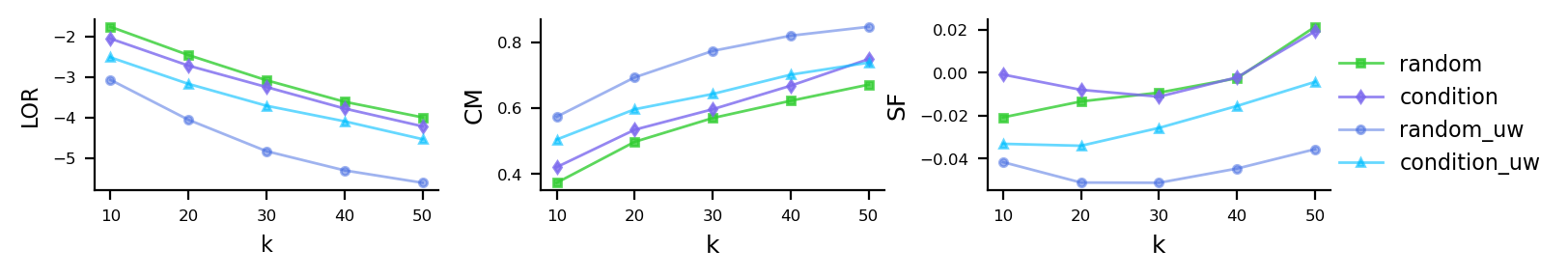}
     \caption{Evaluation performance over BERT on 20Newsgroup.}
     \label{fig2}
 \end{subfigure}
 \caption{Evaluation performances on the different datasets over BERT architecture.}
 \label{err}
\end{figure*}

Fig.\ref{err} demonstrates the evaluation performance over BERT architecture on different datasets with deletion operation. We can observe that \textbf{random-uw} outperforms other baselines. \textbf{random-uw} and \textbf{condition-uw} always perform better than the corresponding baselines without uncertainty-based reweighting. In SST-2, SNLI and Yelp-2, random-based baselines obtain better results compared with condition-based baselines while for SNIPS and 20Newsgroup \textbf{condition-uw} can outperform \textbf{random}.

\section{Performance over RoBERTa architecture}
\label{appendixro}

\begin{figure*}
 \centering
 \begin{subfigure}[b]{\textwidth}
     \centering
     \includegraphics[width=\textwidth,height=1in]{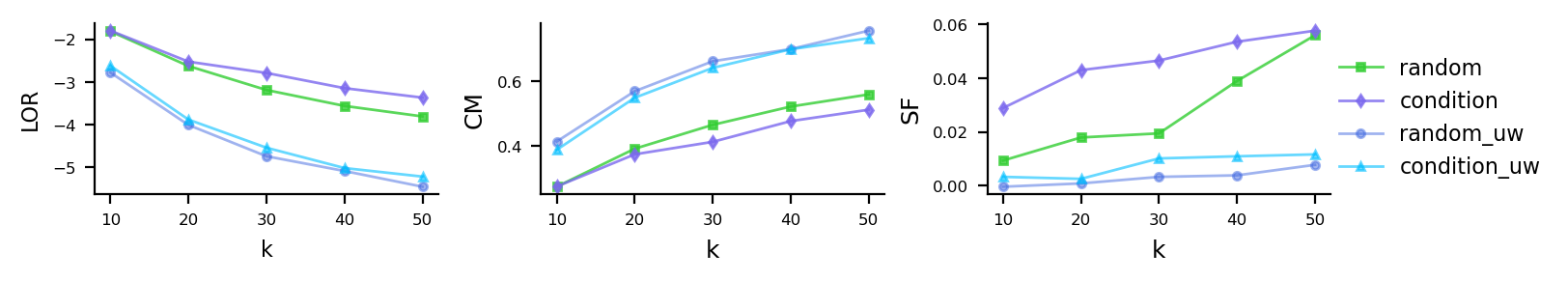}
     \caption{Evaluation performance over RoBERTa on SST-2.}
     \label{fig1}
 \end{subfigure}
 \begin{subfigure}[b]{\textwidth}
     \centering
     \includegraphics[width=\textwidth,height=1in]{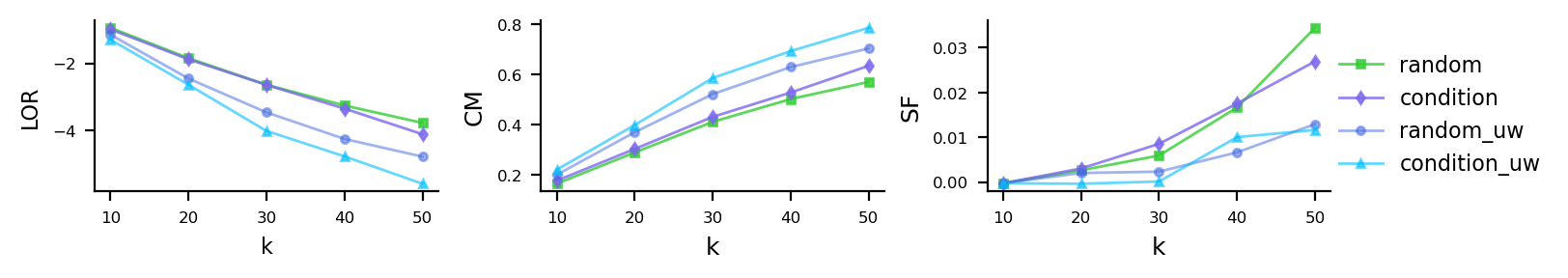}
     \caption{Evaluation performance over RoBERTa on SNIPS.}
     \label{fig2}
 \end{subfigure}
 \begin{subfigure}[b]{\textwidth}
     \centering
     \includegraphics[width=\textwidth,height=1in]{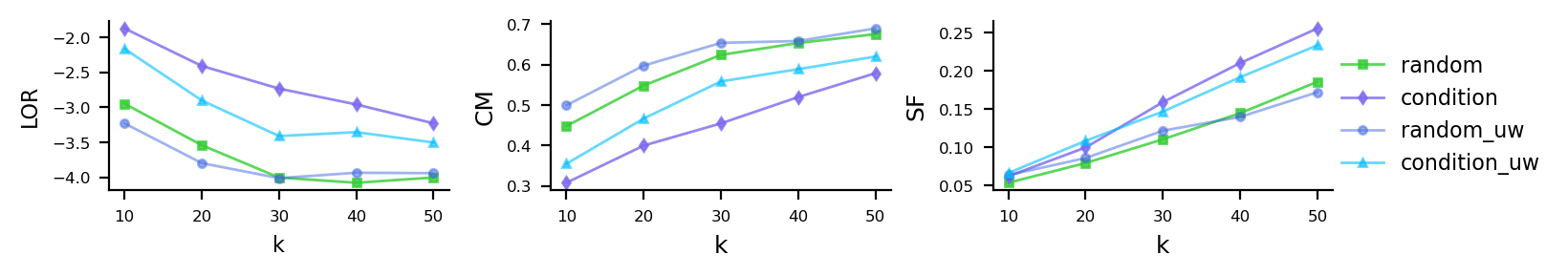}
     \caption{Evaluation performance over RoBERTa on SNLI.}
     \label{fig3}
 \end{subfigure}
 \begin{subfigure}[b]{\textwidth}
     \centering
     \includegraphics[width=\textwidth,height=1in]{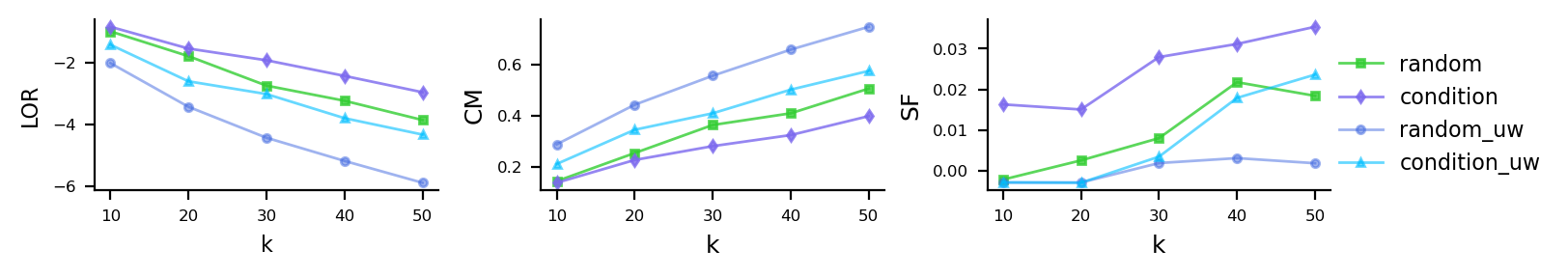}
     \caption{Evaluation performance over RoBERTa on Yelp-2.}
     \label{fig4}
 \end{subfigure}
 \begin{subfigure}[b]{\textwidth}
     \centering
     \includegraphics[width=\textwidth,height=1in]{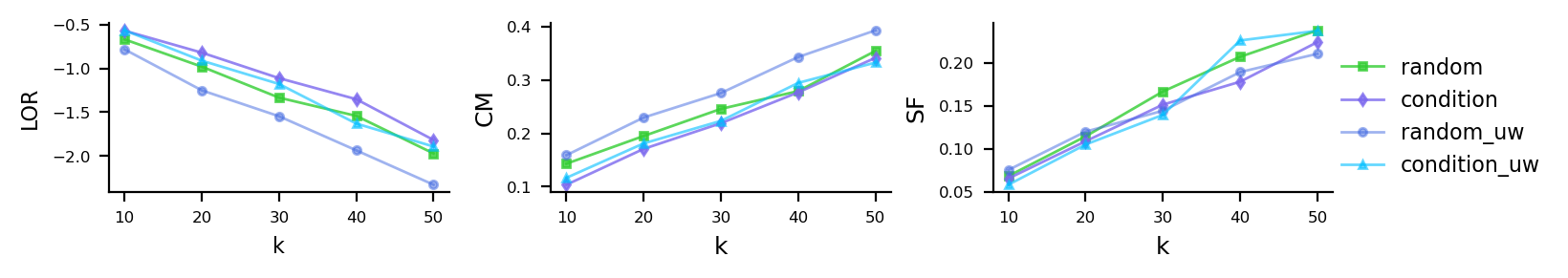}
     \caption{Evaluation performance over RoBERTa on 20Newsgroup.}
     \label{fig2}
 \end{subfigure}
 \caption{Evaluation performances on the different datasets over RoBERTa architecture.}
 \label{ro}
\end{figure*}

Padding and deletion operations generate the same results over RoBERTa architecture. RoBERTa's training process might make it more robust to variations in input. We can read from Fig.\ref{ro} that \textbf{random-uw} and \textbf{condition-uw} achieve comparable performance on SST-2 and SNIPS. Introducing uncertainty-reweighting mechanism can help improve the faithfulness of Shapley explanation.

\section{Robustness of Shapley Sampling method}
We choose smaller sample size $100$ with $5$ different runs. Table \ref{tab:tin_sst2}, \ref{tab:tin_snips} and \ref{tab:tin_20newsgroup} report the average performance on SST-2, SNIPS and 20Newsgroup over BERT architecture. We can observe that Shapley Sampling method is relatively robust and uncertainty-based reweighting mechanism does increase the performance.

\label{rob}
\begin{table}[H]
\centering
\caption{Evaluation performance over BERT on SST-2 with $k=20$.}
\vskip 0.15in
\label{tab:tin_sst2}
\begin{tabular}{lccl} 
\hline
\multirow{2}{*}{Methods} & \multicolumn{3}{c}{SST-2} \\
                         & LOR & CM & SF \\
\hline
random                   & -2.6281$\pm$0.0720 & 0.3884$\pm$0.0082 & 0.0021$\pm$0.0010 \\
condition                & -2.2590$\pm$0.0608 & 0.3515$\pm$0.0082 & 0.0272$\pm$0.0020 \\
random\_uw               & -3.9181$\pm$0.0661 & 0.5533$\pm$0.0067 & -0.0090$\pm$0.0010 \\
condition\_uw            & -3.1907$\pm$0.0600 & 0.4635$\pm$0.0065 & 0.0014$\pm$0.0031 \\
\hline
\end{tabular}
\end{table}

\begin{table}[H]
\centering
\caption{Evaluation performance over BERT on SNIPS with $k=20$.}
\vskip 0.15in
\label{tab:tin_snips}
\begin{tabular}{lccl} 
\hline
\multirow{2}{*}{Methods} & \multicolumn{3}{c}{SNIPS} \\
                         & LOR & CM & SF \\
\hline
random                   & -1.1928$\pm$0.0369 & 0.2436$\pm$0.0065 & 0.0029$\pm$0.0023 \\
condition                & -1.1288$\pm$0.0432 & 0.2322$\pm$0.0074 & 0.0061$\pm$0.0008 \\
random\_uw               & -1.7017$\pm$0.0447 & 0.3317$\pm$0.0074 & -0.002$\pm$0.0015 \\
condition\_uw            & -1.5808$\pm$0.0301 & 0.3100$\pm$0.0060 & 0.0008$\pm$0.0014 \\
\hline
\end{tabular}
\end{table}

\begin{table}
\centering
\caption{Evaluation performance over BERT on 20Newsgroup with $k=20$.}
\vskip 0.15in
\label{tab:tin_20newsgroup}
\begin{tabular}{lccl} 
\hline
\multirow{2}{*}{Methods} & \multicolumn{3}{c}{20Newsgroup} \\
                         & LOR & CM & SF \\
\hline
random                   & -2.4960$\pm$0.0638 & 0.4706$\pm$0.0062 & -0.0258$\pm$0.0048 \\
condition                & -2.7000$\pm$0.1045 & 0.5301$\pm$0.0181 & -0.0186$\pm$0.0076 \\
random\_uw               & -4.1049$\pm$0.0646 & 0.7014$\pm$0.0118 & -0.0567$\pm$0.0032 \\
condition\_uw            & -3.2504$\pm$0.0852 & 0.5952$\pm$0.0152 & -0.0356$\pm$0.0046 \\
\hline
\end{tabular}
\end{table}

\section{In-distribution baselines}
\begin{figure}
 \centering
 \begin{subfigure}[b]{0.25\textwidth}
     \centering
     \includegraphics[width=\textwidth]{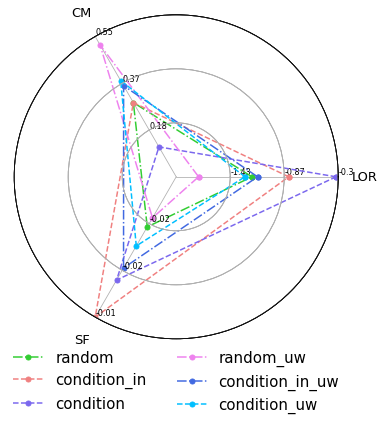}
     \caption{Yelp-2.}
     \label{si}
 \end{subfigure}
 \begin{subfigure}[b]{0.25\textwidth}
     \centering
     \includegraphics[width=\textwidth]{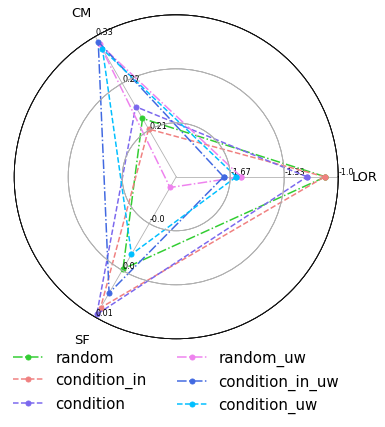}
     \caption{SNIPS.}
     \label{ag}
 \end{subfigure}
 \caption{Evaluation performance with $k=20$ in Yelp-2 and SNIPS.}
  \label{indi2}
\end{figure}

Fig.\ref{indi2} shows the evaluation performance with $k=20$ on Yelp-2 and SNIPS. Similarly, in-distribution baselines outperforms their corresponding baselines without in-distribution sampling.
\label{indii}

\section{Human evaluation}
\label{he}
\begin{figure}
 \centering
 \begin{subfigure}[b]{0.5\textwidth}
     \centering
     \includegraphics[width=\textwidth,height=1in]{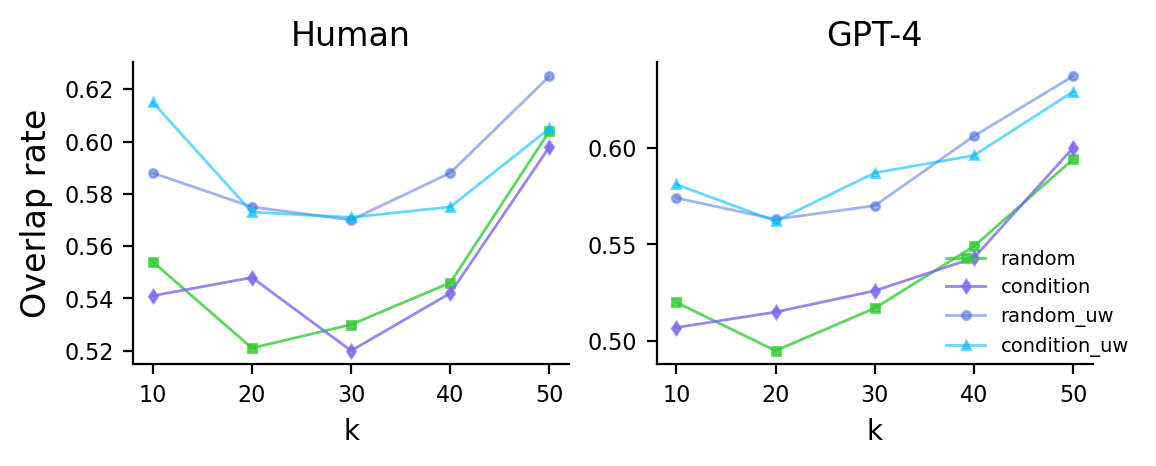}
     \caption{SST-2}
     \label{si}
 \end{subfigure}
 \begin{subfigure}[b]{0.5\textwidth}
     \centering
     \includegraphics[width=\textwidth,height=1in]{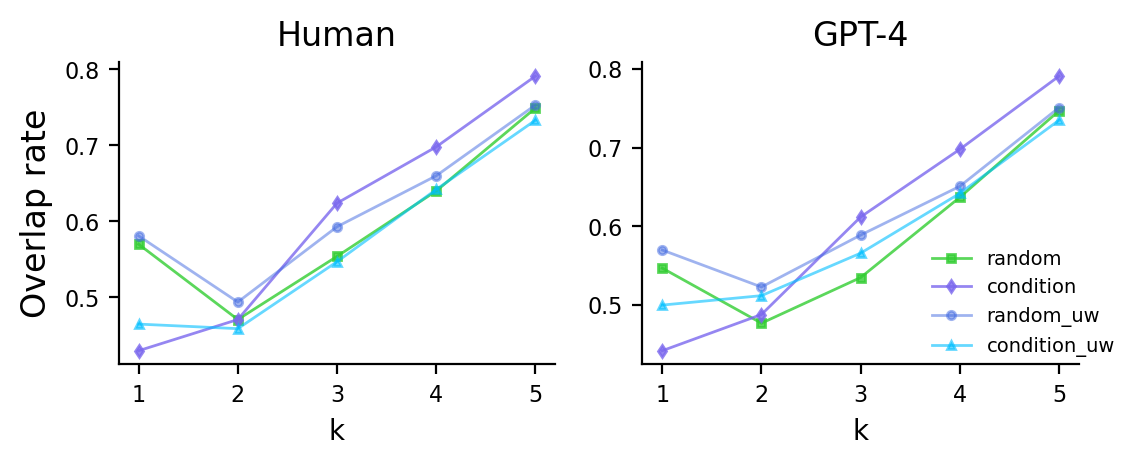}
     \caption{SNIPS}
     \label{ag}
 \end{subfigure}
 \caption{Overlap rate of top influential features between GPT-4 (or human) and baselines over RoBERTa architecture in SST-2 and SNIPS.}
  \label{indi33}
\end{figure}

\begin{figure*}[h]
 \centering
 \begin{subfigure}[b]{\textwidth}
     \centering
     \includegraphics[width=\textwidth,height=1in]{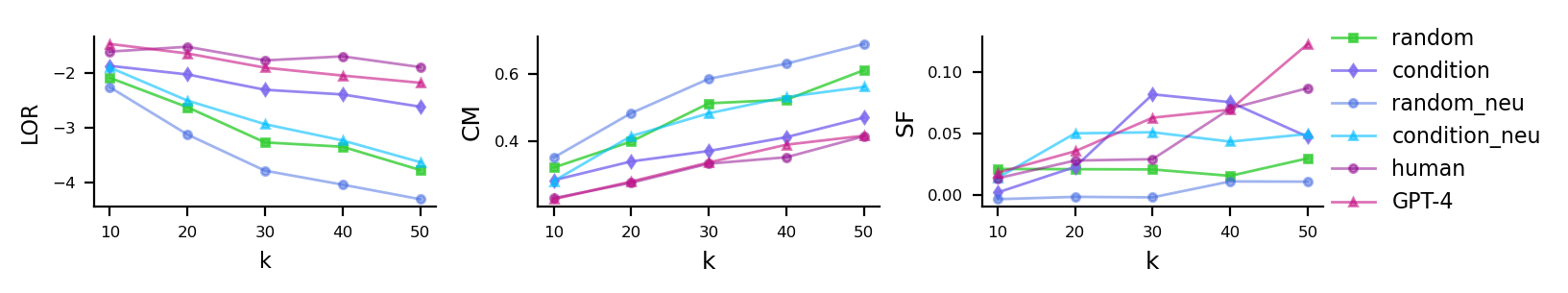}
     \caption{Evaluation performance over BERT on SST-2.}
     \label{fig1}
 \end{subfigure}
 \begin{subfigure}[b]{\textwidth}
     \centering
     \includegraphics[width=\textwidth,height=1in]{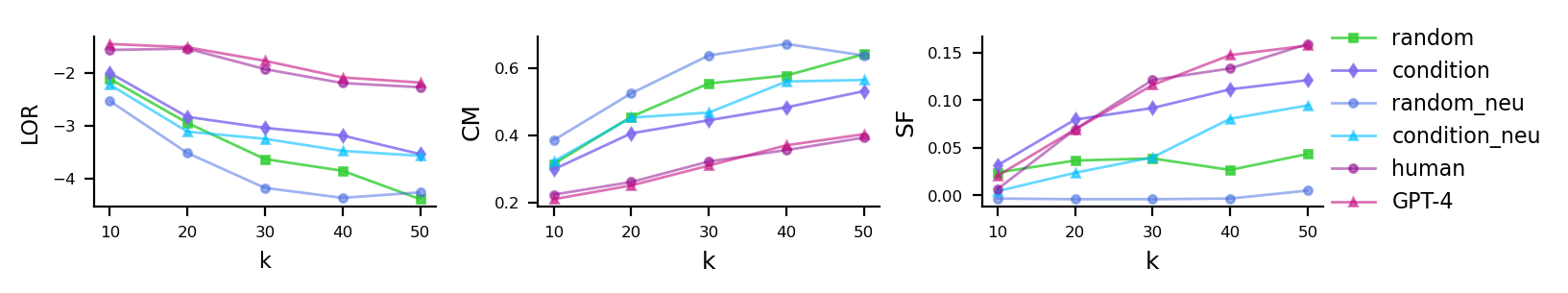}
     \caption{Evaluation performance over RoBERTa on SST-2.}
     \label{fig2}
 \end{subfigure}
 \begin{subfigure}[b]{\textwidth}
     \centering
     \includegraphics[width=\textwidth,height=1in]{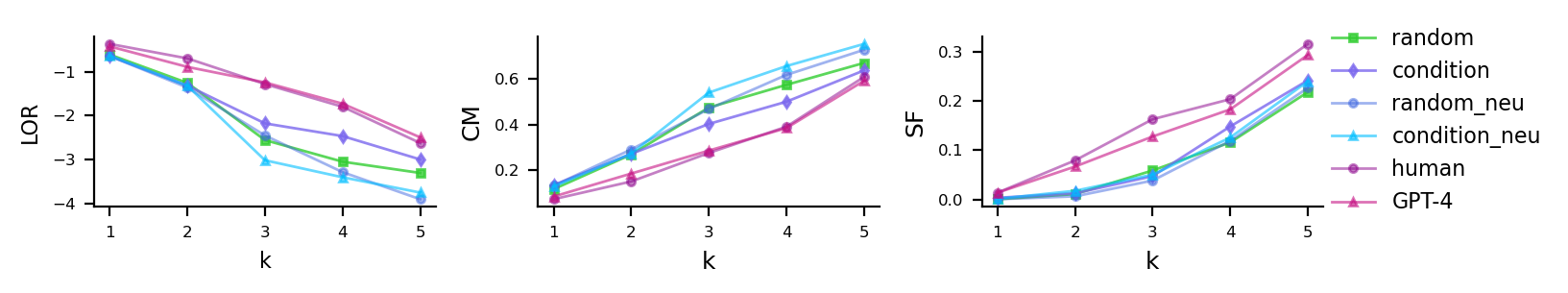}
     \caption{Evaluation performance over BERT on SNIPS.}
     \label{fig3}
 \end{subfigure}
 \begin{subfigure}[b]{\textwidth}
     \centering
     \includegraphics[width=\textwidth,height=1in]{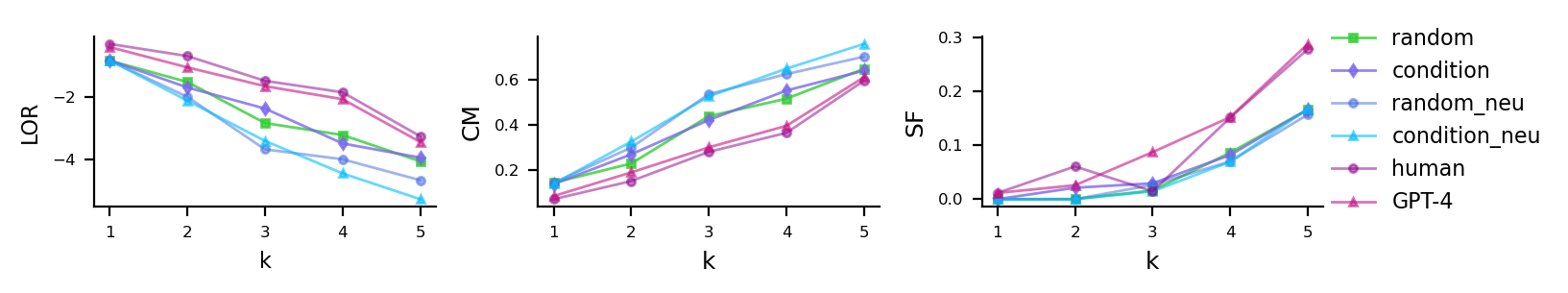}
     \caption{Evaluation performance over RoBERTa on SNIPS.}
     \label{fig4}
 \end{subfigure}
 \caption{Evaluation performances on SST-2 and SNIPS.}
 \label{hgss}
\end{figure*}

Table \ref{t2} demonstrates how to use GPT-4 to generate explanations and Table \ref{th1} shows an example of explanations generated by different baselines, GPT-4 and human over BERT architecture. Table \ref{conro} and Fig.\ref{indi33} shows that the explanations provided by GPT-4 (or human) are not closely correlated with the explanation generated by different baselines. Fig.\ref{hgss} reports the quantitative analysis of explanations provided by GPT-4 and human.

\begin{table*}[]
\caption{Illustration of how to use GPT-4 to generate explanations. ``Input'' and ``Output'' refer to the prompt provided to GPT-4 and the generated explanations, respectively. It could be treated as zero-shot evaluation. We maintained the output integrity without alternations, while occasionally adjusting the requirements to ensure a complete ranking. For instance, when dealing with repeated strings, each instance was assigned an individual rank.}
\vskip 0.15in
\centering
\label{t2}
\begin{tabularx}{\textwidth}{|c|X|}
\hline
\textbf{Component} & \textbf{Description} \\ \hline
\multirow{8}{*}{Input} & The task is described as follows: given a text sequence of a movie review with the sentiment classification label (positive or negative), there are a few requirements:
\begin{enumerate}
    \item Transform this long string sequence into a list of strings (denoted as a transformed list).
    \item Measure the contributions of each string in the list toward the sentiment label based on your understanding. Then rank all strings (ensuring that no strings are excluded) including the repeated strings (each occurrence should have its own rank) in this list based on their contributions.
    \item The ranking should follow an order from the most positive to neutral to the most negative. Place the strings with the highest positive contribution at the top and the strings with the most negative contribution at the bottom.
    \item Output all ranked strings ensuring that no strings are excluded.
\end{enumerate} \\ \hline
\multirow{2}{*}{Example Input} & Sequence: something the true film buff will enjoy \newline
Label: positive \\ \hline
\multirow{2}{*}{Example Output} & Transformed list: [`something', `the', `true', `film', `buff', `will', `enjoy']\newline
Ranked strings: [`enjoy', `true', `something', `film', `buff', `will', `the'] \\ \hline
\end{tabularx}
\end{table*}

\begin{table*}
\caption{An example of explanations provided by different baselines, GPT-4 and human over BERT architecture.}  
\vskip 0.15in
\centering
\begin{tabularx}{0.6\textwidth}{c|X}
\toprule
		\hline
\multirow{2}{*}{\textbf{Methods}}&\textbf{sequence}: most new movies have a bright sheen \newline
\textbf{label}: positive \\
\hline  
\multirow{1}{*}{\textbf{random}} 
&`bright', `sheen', `have', `new', `a', `movies', `most'\\
\hline
\multirow{1}{*}{\textbf{condition}} 
& `bright', `movies', `new', `most', `have', `sheen', `a'\\
\hline
\multirow{1}{*}{\textbf{random\_uw}} 
& `bright', `sheen', `have', `new', `a', `movies', `most'\\
\hline
\multirow{1}{*}{\textbf{condition\_uw}} 
&`bright', `movies', `new', `most', `sheen', `have',  `a'\\
\hline
\multirow{1}{*}{\textbf{GPT-4}}& `bright', `sheen', `most', `new', `movies', `have', `a'\\
\hline
\multirow{1}{*}{\textbf{Human}} 
&`bright', `sheen', `most', `new', `movies', `have', `a'\\ 
\hline
\end{tabularx}
\label{th1}
\end{table*}

\begin{table}
 \caption{Rank correlation coefficient between GPT-4 (or human) and baselines over RoBERTa architecture.}
 \vskip 0.15in
\centering
\label{conro}
\begin{tabular}{lcccc} 
\toprule
\multirow{2}{*}{Methods} & \multicolumn{2}{c}{SST-2}           & \multicolumn{2}{c}{SNIPS}            \\
                         & GPT-4            & Human            & GPT-4            & Human             \\ 
\hline
random                   & 0.033            & 0.095            & 0.186            & 0.129             \\
condition                & 0.142            & 0.173            & 0.050            & -0.001            \\
random\_uw               & 0.152            & 0.113            & 0.189            & 0.121             \\
condition\_uw            & 0.203            & 0.137            & 0.094            & 0.038             \\
GPT-4                    & \textbackslash{} & 0.772            & \textbackslash{} & 0.830             \\
Human                    & 0.772            & \textbackslash{} & 0.830            & \textbackslash{}  \\
\bottomrule
\end{tabular}
\end{table}

\begin{figure}
 \centering
 \begin{subfigure}[b]{0.5\textwidth}
     \centering
     \includegraphics[width=\textwidth,height=1in]{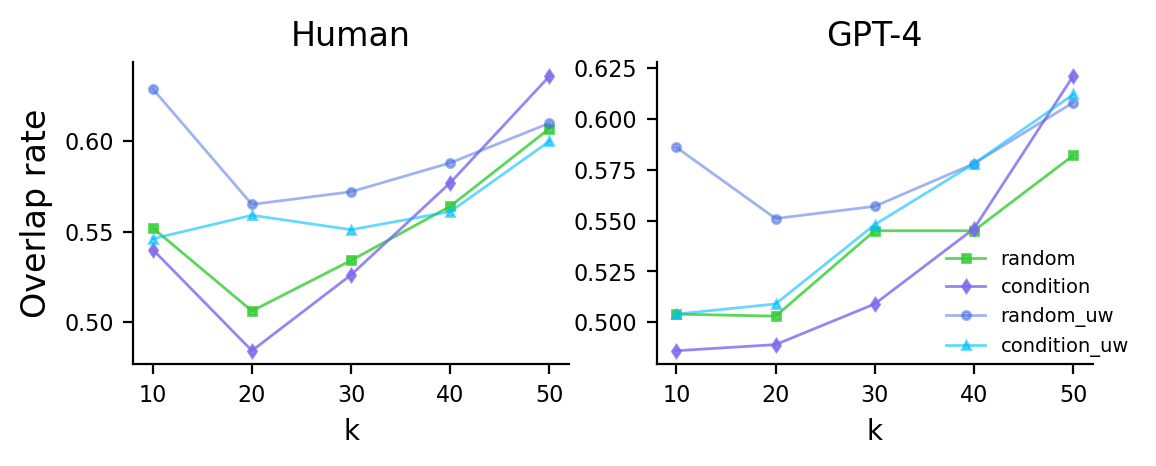}
     \caption{SST-2.}
     \label{si}
 \end{subfigure}
 \begin{subfigure}[b]{0.5\textwidth}
     \centering
     \includegraphics[width=\textwidth,height=1in]{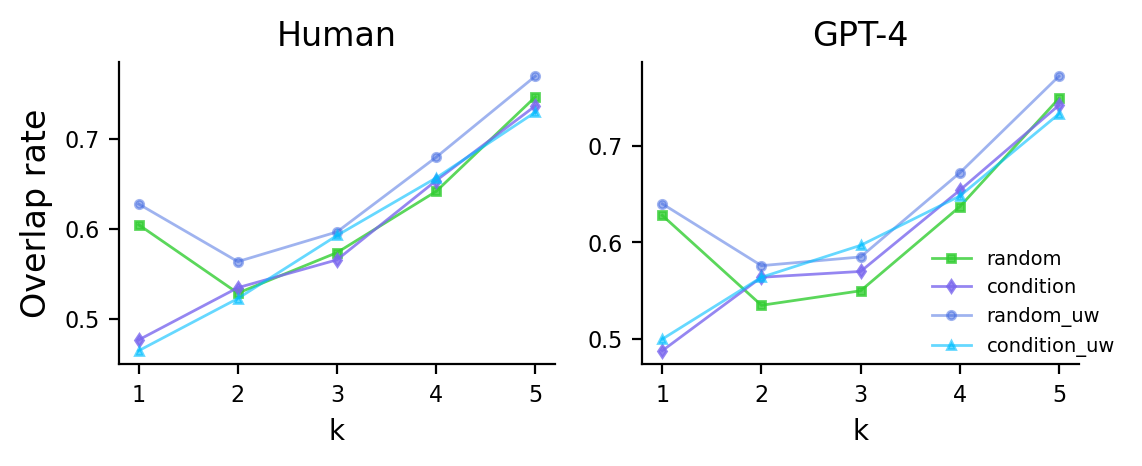}
     \caption{SNIPS.}
     \label{ag}
 \end{subfigure}
 \caption{Overlap rate of top $k$ influential features between GPT-4 (or human) and baselines over BERT architecture on SNIPS.}
  \label{indi3}
\end{figure}

\end{document}